\documentclass[sigconf]{acmart}

\def\BibTeX{{\rm B\kern-.05em{\sc i\kern-.025em b}\kern-.08emT\kern-.1667em\lower.7ex\hbox{E}\kern-.125emX}}
    
\usepackage{xspace}

\newcommand{\combo}{\texttt{Combo}\xspace}
\newcommand{\nt}{\texttt{NT3}\xspace}
\newcommand{\uno}{\texttt{Uno}\xspace}

\newcommand{\async}{\texttt{A3C}\xspace}
\newcommand{\sync}{\texttt{A2C}\xspace}
\newcommand{\rdm}{\texttt{RDM}\xspace}

\usepackage{comment}
\usepackage{enumitem}
\usepackage{graphicx}
\usepackage{subcaption} %
\usepackage{multicol} %
\usepackage{verbatim}

\renewcommand\footnotetextcopyrightpermission[1]{} %
\begin{document}

\title[RL-based NAS for cancer DL research]{Scalable Reinforcement-Learning-Based Neural Architecture Search for Cancer Deep Learning Research}

\author{Prasanna Balaprakash}
\authornote{Both authors contributed equally to this research.}
\email{pbalapra@anl.gov}
\affiliation{%
 \institution{Argonne National Laboratory}
}
\author{Romain Egele}
\authornotemark[1]
\email{regele@anl.gov}
\affiliation{%
 \institution{Argonne National Laboratory}
}
\author{Misha Salim}
\email{msalim@anl.gov}
\affiliation{%
    \institution{Argonne National Laboratory}
}
\author{Stefan Wild}
\email{wild@anl.gov}
\affiliation{%
    \institution{Argonne National Laboratory}
}
\author{Venkatram Vishwanath}
\email{venkat@anl.gov}
\affiliation{
    \institution{Argonne National Laboratory}
}
\author{Fangfang Xia}
\email{fangfang@anl.gov}
\affiliation{
    \institution{Argonne National Laboratory}
}
\author{Tom Brettin}
\email{brettin@anl.gov}
\affiliation{
    \institution{Argonne National Laboratory}
}
\author{Rick Stevens}
\email{stevens@anl.gov}
\affiliation{
    \institution{Argonne National Laboratory}
}

\renewcommand{\shortauthors}{Balaprakash and Egele, et al.}

\begin{abstract}
Cancer is a complex disease, the understanding and treatment of which are being aided through increases in the volume of collected data and in the scale of deployed computing power. 
Consequently, there is a growing need for the development of data-driven and, in particular, deep learning methods for various tasks such as cancer diagnosis, detection, prognosis, and prediction. Despite recent successes, however,  designing high-performing deep learning models for nonimage and nontext cancer data is a time-consuming, trial-and-error, manual task that requires both cancer domain and deep learning  expertise. To that end, we develop a reinforcement-learning-based neural architecture search to automate deep-learning-based predictive model development for a class of representative cancer data. We develop custom building blocks that allow domain experts to incorporate the cancer-data-specific characteristics. We show that our approach discovers deep neural network architectures that have significantly fewer trainable parameters, shorter training time, and accuracy similar to or higher  than those of manually designed architectures. We study and demonstrate the scalability of our approach on up to 1,024 Intel Knights Landing nodes of the Theta supercomputer at the Argonne Leadership Computing Facility.  
\end{abstract}


\copyrightyear{2019} 
\acmYear{2019} 

\maketitle
\section{Introduction}

Cancer is a disease that drastically alters the normal biological function of cells and damages the health of an individual. 
Cancer is estimated to be 
the second leading cause of death globally and was responsible for 9.6 million deaths in 2018 \cite{CancerStat}.  
A thorough understanding of cancer remains elusive because of challenges due to the variety of cancer types,  heterogeneity within a cancer type, structural variation in cancer-causing genes, complex metabolic pathways, and nontrivial drug-tumor interactions \cite{ling2015extremely,nikolaou2018challenge,reznik2018landscape,dixon2018integrative,sanchez2018oncogenic}. 

Recently, as a result of coordinated data management initiatives, the cancer research community increasingly has access to a large volume of data. This has led to a number of promising large-scale, data-driven cancer research efforts. %
In particular, machine learning (ML) methods have been employed for tasks such as identifying cancer cell patterns; modeling complex relationships between drugs and cancer cells; and predicting cancer types.

With the sharp increases in available data and computing power, considerable attention has been devoted to deep learning (DL) approaches. %
The first wave of success in applying DL for cancer stems from adapting the convolutional neural network (CNN) and recurrent neural network (RNN) architectures that were developed for image and text data. For example, CNNs have been used for cancer cell detection from images, and RNNs and its variants have been used for analyzing  clinical reports. 

These adaptations are possible because of the underlying regular grid nature of image and text data \cite{bronstein2017geometric}. For example, images share the spatial correlation properties, and convolution operations designed to extract features from natural images can be generalized for detecting cancer cells in images with relatively minor modifications. 
However, designing deep neural networks (DNNs) for nonimage and nontext data remains underdeveloped in cancer research. Several cancer predictive modeling tasks deal with \emph{tabular data} comprising an output and multidimensional inputs. For example, in the drug response problem, DNNs can be used to model a complex nonlinear relationship between the properties of drugs and tumors in order to predict treatment response \cite{xia2018predicting}. Here, the properties of drugs and tumors cannot easily be expressed as images and text and cast into classical CNN and RNN architectures. Consequently, cancer researchers and DL experts resort to manual trial-and-error methods to design DNNs. Tabular data types are diverse;  consequently, designing DNNs with shared patterns such as CNNs and RNNs is not meaningful unless further assumptions about the data are made. Fully connected DNNs are used for many modeling tasks with tabular data. However, they can potentially lead to unsatisfactory performance because they can have large numbers of parameters, overfitting issues, and difficult optimization landscape with low-performing local optima \cite{fernandez2014we}. Moreover, tabular data often is obtained from multiple sources and modes, where combining certain inputs using \emph{problem-specific} domain knowledge can lead to better features and physically meaningful and robust models, thus preventing the design of effective architectures similar to CNNs and RNNs.

Automated machine learning (AutoML) \cite{automl,automl-google,bergstra2013making,AutoMLBook2019,zoph2016neural} automates the development of ML models by searching over appropriate components and their hyperparameters for preprocessing, feature engineering, and model selection to maximize a user-defined metric. AutoML has been shown to reduce the amount of human effort and time required for a number of traditional ML model development tasks. Although DL reduces the need for feature engineering, extraction, and selection tasks, finding the right DNN architecture and its hyperparameters is crucial for predictive accuracy. Even on  image and text data, DNNs obtained by using AutoML approaches have outperformed manually engineered DNNs that took several years of development \cite{zoph2016neural,young2017evolving,patton2018167}. 

AutoML approaches for DNNs can be broadly classified into hyperparameter search and neural architecture search (NAS). 
Hyperparameter search approaches try to find the best values for the hyperparameters for a fixed neural architecture.
Examples include random search \cite{bergstra2012random}, Bayesian optimization 
\cite{snoek2012practical,bergstra2013hyperopt,klein-bayesopt17}, 
bandit-based methods \cite{li2016hyperband,snoek2012practical}, 
metaheuristics \cite{lorenzo2017hyper,miikkulainen2017evolving}, and 
population-based training \cite{jaderberg2017population} approaches.
NAS methods search over model descriptions of neural network specifications. Examples include discrete search-space traversal
\cite{negrinho2017deeparchitect,liu2017progressive}, 
reinforcement learning (RL) 
\cite{baker2016designing,zoph2016neural,pham2018efficient}, and
evolutionary algorithms \cite{floreano2008neuroevolution,stanley2009hypercube,suganuma2017genetic,wierstra2005modeling}. 

We focus on developing scalable neural-network-based RL for NAS, which offers several potential advantages. 
First, RL is a first-order method that leverages gradients. Second, RL-based NAS construction is based on a Markov decision process: decisions that are made to construct a given layer depend on the decisions that were made on the previous layers.  This exploits the inherent structure of DNNs, which are characterized by hierarchical data flow computations; 
While traditional RL methods pose several challenges \cite{sutton2018reinforcement} such as exploration-exploitation tradeoff, sample inefficiency, and long-term credit assignment, recent developments \cite{schulman2017proximal,grondman2012survey} in the field are promising.

Although hyperparameter search work has been done on cancer data \cite{WozBMC18}, to our knowledge scalable RL-based NAS has not been applied to cancer predictive modeling tasks. An online bibliography of NAS \cite{nasonlinebib2019} and a recent NAS survey paper \cite{elsken2018neural} did not list any cancer-related articles. The reasons may be twofold. First, AutoML, and in particular NAS, is still in its infancy. Most of the existing work in NAS  focuses on image classification tasks on benchmark data sets. Since convolutions and recurrent cells form the basic building blocks for CNNs and RNNs, respectively, the problem of defining the search space for CNN and RNN architectures has become relatively easy \cite{elsken2018neural}. However,  no such generalized building block exists for nonimage and nontext data. Second, large-scale NAS and hyperparameter search require high-performance computing (HPC) resources and appropriate software infrastructure \cite{distdl-preprint}. These requirements are attributed to the fact that architecture evaluations (training and validation) are computationally expensive and parallel evaluation of multiple architectures on multiple compute nodes through scalable search methods is critical to finding DNNs with high accuracy in short computation time. We note that the time needed for NAS can be more than the training time of a manually designed network. However, designing the network by manually intensive trial-and-error approaches can take days to weeks even for ML experts \cite{AutoMLBook2019}.

We develop a scalable RL-based NAS infrastructure to automate DL-based predictive model development for a class of cancer data. The contributions of the paper are as follows:
\vspace{-0.2cm}
\begin{itemize}[leftmargin=*]
\item We develop a DL NAS search space with new types of components that take into account characteristics specific to cancer data. These include multidrug and cell line inputs, 1D convolution for traversing large drug descriptors, and 
    nodes that facilitate weight sharing between drug descriptors.
    \item We demonstrate a scalable RL-based NAS on 1,024 Intel Knights Landing (KNL) nodes of Theta, a leadership-class HPC system, for cancer DL using a multiagent and multiworker approach.
    \item We scale asynchronous and synchronous proximal policy optimization, a state-of-the-art RL approach for NAS. Of particular importance is the convergence analysis of search methods at scale. We demonstrate that RL-based NAS achieves high accuracy on architectures because of the search strategy and not by pure chance as in random search. 
    \item We show that the scalable RL-based NAS can be used to generate multiple  accurate DNN architectures that have significantly fewer training parameters,  shorter training time, and accuracy similar to or higher  than those of manually designed networks. 
    \item We implement our approach as a neural architecture search module within DeepHyper \cite{deephyper,deephyper-soft2018}, an open-source  software package, that can be readily deployed on leadership-class machines for cancer DL research.
\end{itemize}

\section{Problem sets and  manually designed deep neural networks}

We focus on a set of DL-based predictive modeling problem sets from the CANcer Distributed Learning Environment (CANDLE) project \cite{candle-ecp}  that comprises data sets and manually designed DNNs for drug response; RAS gene family pathways; and treatment strategy at  molecular, cellular, and population scales. Within these problem sets, we target three 
benchmarks, which represent a class of predictive modeling problems that seek to predict drug response based on molecular features of tumor cells and drug descriptors. 
An overview of these open-source \cite{candle-code} benchmarks (i.e., data set plus manually designed DNN) is given below.

\subsection{Predicting tumor cell line response 
(\combo)}
In the \combo benchmark \cite{combo-code}, recent paired drug screening results from the National Cancer Institute (NCI) 
are used to model drug synergy and understand how drug combinations interact with tumor molecular features. Given 
drug screening results on NCI60 cell lines available at the NCI-ALMANAC database, \combo's goal is to build a DNN that can predict the growth percentage from the cell line molecular features and the descriptors of drug pairs. The manually designed DNN comprises three input layers: one for cell expression (of dimension $d=$942) and two for drug descriptors ($d=$3,820). The two input layers for the drug pairs are connected by a shared submodel of three dense layers ($d=$1,000). The cell expression layer is connected to a submodel of three dense layers 
each with $d=$1,000. The outputs of these submodels are concatenated and connected to three dense 
layers each with $d=$1,000. The scalar output layer is used to predict the percent growth for a given drug concentration. The training and validation input data are given as matrices of sizes 248,650 (number of data points) $\times$ 4,762 (total input size) and 62,164 $\times$ 4,762, respectively. The training and validation output data are matrices of size 248,650 $\times$ 1 and 62,164 $\times$ 1, respectively.

\subsection{Predicting tumor dose response across multiple data sources (\uno)}
The \uno benchmark \cite{uno-code} integrates cancer drug screening data from 2.5 million samples across six research centers to examine study biases and to build a unified
drug response model. 
The 
associated manually designed DNN has four  input layers: a cell RNA sequence layer ($d=$942), a dose layer ($d=$1), a drug descriptor layer ($d=$5,270), and a drug fingerprints layer ($d=$2,048).
It has three feature-encoding submodels for cell RNA sequence, drug descriptor, and drug fingerprints. Each submodel is composed of three hidden layers, each with $d=$1,000. The last layer for each of the submodels is connected to the concatenation layer along with the dose layer. This is connected to three hidden layers 
each with $d=$1,000. The scalar output layer is used to predict tumor dose. We used the single drug paclitaxel, a simplified indicator, for this study. The training and validation input data are given as matrices of sizes 9,588 $\times$ 8,261 and 2,397 $\times$ 8,261, respectively. The training and validation output data are given as matrices of sizes 9,588 $\times$ 1 and 2,397 $\times$ 1, respectively.

\subsection{Classifying RNA-seq gene expressions 
(\nt)}
The \nt benchmark  \cite{nt3-code} classifies tumors from normal tissue
by tracking gene-expression-level tumor signatures.
The associated manually designed DNN has an input layer for RNA sequence gene expression ($d$=60,483). This is connected to a 
1D convolutional layer of 128 filters with kernel size 20 and a maximum pooling layer of size 1. This is followed by a 
1D convolutional layer of 128 filters with kernel size 10 and a maximum pooling layer of size 10. The output of the pooling layer is flattened and given to the dense layer of size 200 and a dropout layer with 0.1\%. This is followed by a dense layer of size 20 and a dropout layer with 0.1\%. The output  layer of size 2 for the two classes with softmax activation is used to predict the tissue type.
The training and validation input data are given as matrices of sizes 1,120 $\times$ 60,483 and 280 $\times$ 60,483, respectively. 
The training and validation output data are given as matrices of sizes 1,120 $\times$ 1 and 280 $\times$ 1, respectively.

\section{RL-Based NAS}
\label{rl-nas}
NAS comprises (1) a search space that defines a set of feasible architectures, (2) a search strategy to search over the defined search space, and (3) a reward estimation strategy that describes how to evaluate the quality 
of a given neural architecture.  

\subsection{Search space}

We describe the search space of a neural architecture  using a graph structure. The basic building block is a set of nodes $\mathcal{N}$ with possible choices; typically these choices are nonordinal (i.e., values that cannot be ordered in a numeric scale). For example, \{Dense(10, sig), Dense(50, relu), and Dropout(0.5)\} respectively represent a dense layer with 10 units and sigmoid activation, a dense layer with 50 units with relu activation, and a layer with 50\% dropout. A block $B$ is a directed acyclic graph: $(\mathcal{N}=(\mathcal{N}^I, \mathcal{N}^O, \mathcal{N}^R), \mathcal{R}_{\mathcal{N}})$, where the set of nodes is differentiated by  input nodes $\mathcal{N}^I$, intermediate nodes $\mathcal{N}^R$, and output nodes $\mathcal{N}^O$ and where  $\mathcal{R}_{\mathcal{N}} \subseteq (\mathcal{N}^I \cup \mathcal{N}^R) \times (\mathcal{N}^R \cup \mathcal{N}^O)$ is a set of binary relations\footnote{Without loss of generality, this can be extended to multiple intermediate nodes.}  that describe the 
connections among 
nodes in $\mathcal{N}$
A cell $C_i$ consists of a set of $L_i$
blocks $\{B_i^0, ..., B_i^{L_i-1}\}$ and a rule $R_{C_{\rm out}}$ 
to create the output of $C_i$. The structure $S$ is given by the set $\{(I_S^0, \ldots,I_S^{P-1}), (C_0, \ldots, C_{K-1}), R_{S_{\rm out}}\}$, where $(I_S^0, \ldots,I_S^{P-1})$ is a tuple of $P$ 
inputs, $(C_0, \ldots, C_{K-1})$ is a tuple of $K$ cells, and $R_{S_{\rm out}}$ is a rule to create the output of $S$. Users can define cell-specific blocks and block-specific input, intermediate, and output nodes.

\begin{figure}
    \centering
    \includegraphics[width=0.45\textwidth]{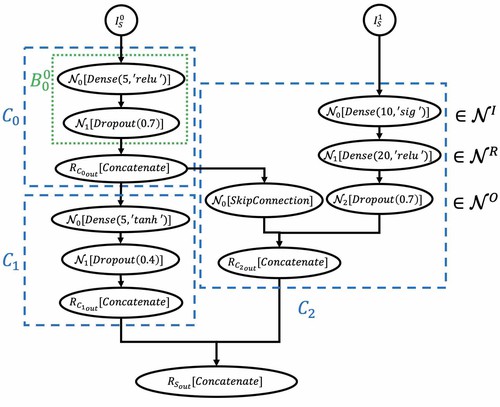}
    \caption{Example search space for NAS}
    \label{fig:formal_structure}
\end{figure}

Figure~\ref{fig:formal_structure} shows a sample search space. The structure $S$ is made up of three cells ($C_0$, $C_1$, $C2$). Cell $C_0$ has one block $B_0$ that has one input ($\mathcal{N}_0$) and one output node ($\mathcal{N}_1$). The rule to create the output is concatenation. Since there is only one block, the output from $\mathcal{N}_1$ is the output layer for $C_0$; cell $C_1$ is similar to cell $C_0$. Since  $\mathcal{N}_0$ is a dense layer, the output of $C_0$ is connected as an input to the $C_1$.

The search space definition that we have  differs in two ways from existing chain-structured neural networks, multibranch networks, and cells blocks for designing CNNs and RNNs. The first is the flexibility to define multiple input layers ($I_{S}^{0},\dots,I_{S}^{J-1}$)
(e.g., to support cell expression, drug descriptors in \combo; and RNA sequence, dose, drug descriptor, and drug fingerprints in \uno) and a cell for each of them. The second is the node types. By default, each node is a \texttt{VariableNode}, which is characterized by a set of possible choices. In addition, we define two types of nodes. \texttt{ConstantNode}, with a particular operation, is excluded from the search space but will be used in neural architecture construction. This allows for domain knowledge encoding---for example, if we want the dose value in \uno in every block, we can define a constant node for every block and connect them to the dose input layer. \texttt{MirrorNode} is used to reuse an existing node. For example, in \combo, \texttt{drug1.descriptors} and \texttt{drug2.descriptors} share the same submodel for feature encoding. To support such shared submodel construction, we define a cell with variable nodes for \texttt{drug1.descriptors} and a cell with mirror nodes \texttt{drug2.descriptors}. Consequently, the mirror nodes are not part of the specified search space.

Using the search space formalism, we define the search spaces for \uno, \combo, and \nt. We consider  a small and a large search space for each of \combo and \uno. For \nt, we define only a small search space because the baseline DNN obtains 98\% accuracy on the validation data.

\subsubsection{\combo} 
We define a \texttt{VariableNode} as a node consisting of  options representing
identity operation,
a dense layer with x units and activation function y (Dense(x, y) ), 
and a dropout layer Dropout($r$) where $r$ is a fraction of input units to drop (e.g., Identity, Dense(100, $relu$), Dense(100, $tanh$), Dense(100, $sigmoid$), Dropout(0.05),\\ Dense(500, $relu$), Dense(500, $tanh$), Dense(500, $sigmoid$), Dropout(0.1),\\ Dense(1000, $relu$), Dense(1000, $tanh$), Dense(1000, $sigmoid$),\\  Dropout(0.2)).
We refer to this \texttt{VariableNode} as \texttt{MLP\_Node}, where \texttt{MLP} stands for multilayered perceptron. 

For the small \combo search space, we define cells $C_0$, $C_1$, and $C_2$. Cell $C_0$ receives input from three input 
layers, \texttt{cell expression}, \texttt{drug 1 descriptors}, and \texttt{drug 2 descriptors}, and has three blocks, $\left<C_0, B_0\right>$, $\left<C_0, B_1\right>$, and $\left<C_0, B_2\right>$. The block $\left<C_0, B_0\right>$ receives input from the \texttt{cell expression} layer and comprises three \texttt{MLP\_Node}s connected sequentially in a feed-forward manner. The block $\left<C_0, B_1\right>$ receives input from \texttt{drug 1 descriptors} and has three \texttt{MLP\_Node}s similar to $\left<C_0, B_0\right>$. The block $\left<C_0, B_2\right>$ receives input from \texttt{drug 2 descriptors} but has three \texttt{Mirror\_Node}s that reuse the \texttt{MLP\_Node}s of $\left<C_0, B_1\right>$ to enable sharing the same submodel between \texttt{drug 1 descriptors} and \texttt{drug 2 descriptors}. The output from $C_0$ is used as input to the cell $C_1$ that contains two blocks, $\left<C_1, B_0\right>$, $\left<C_1, B_1\right>$. The former has three \texttt{MLP\_Node}s with feed-forward connectivity. $\left<C_1, B_1\right>$ has one \texttt{VariableNode} with a \texttt{Connect} operation that includes options to create skip-connections (i.e., Null, Cell expression, Drug 1 descriptors, Drug 2 descriptors, Cell 1 output, Inputs, Cell expression \& Drug 1 descriptors, Cell expression \& Drug 2 descriptors, Drug 1 \& 2 descriptors).

The output from $C_1$ is used as input to the cell $C_2$ that has one block $\left<C_2,B_0\right>$ with three \texttt{MLP\_Nodes} with feed-forward connectivity. The \texttt{Concatenate} operation is used to combine the outputs from $C_0$, $C_1$, and $C_2$ to form the final output. The size of the architecture space is $\approx 2.0968 \times  10^{14}$.

For the large search space, we replicate $C_1$ 8 times. For $C_i$ for $i \in [1,\ldots, 8]$,  we update the set of \texttt{Connect} operations by adding outputs of $C_1, \ldots C_{i-1}$ (i.e., outputs of previous cells). The size of the architecture space is $\approx 2.987 \times 10^{44}$.

\subsubsection{\uno}

For the small search space, we define two cells, $C_0$ and $C_1$. The cell $C_0$ has four blocks, $\left<C_0,B_0\right>$, $\left<C_0,B_1\right>$, $\left<C_0,B_2\right>$, and $\left<C_0,B_3\right>$, that take \texttt{cell rna-seq}, \texttt{dose}, \texttt{drug descriptors}, and \texttt{drug fingerprints} as input, respectively. Each block has three \texttt{MLP\_Nodes} that are connected sequentially. The output rule of $C_0$ is \texttt{Concatenate}. The cell $C_1$ has one block $\left<C_1,B_0\right>$ that takes the $C_0$ output as input and has five nodes: 
$\left<C_1,B_0,\mathcal{N}_0\right>$, 
$\left<C_1,B_0,\mathcal{N}_1\right>$, 
$\left<C_1,B_0,\mathcal{N}_2\right>$, 
$\left<C_1,B_0,\mathcal{N}_3\right>$, and 
$\left<C_1,B_0,\mathcal{N}_4\right>$. 
The nodes $\left<C_1,B_0,\mathcal{N}_2\right>$ and $\left<C_1,B_0,\mathcal{N}_4\right>$ are \texttt{ConstantNode}s with the operation \texttt{Add} (i.e., elementwise addition for tensors). The other three are sequential \texttt{MLP\_Nodes}. The five nodes are connected sequentially, and $\left<C_1,B_0,\mathcal{N}_0\right>$ and $\left<C_1,B_0,\mathcal{N}_2\right>$ are connected to  $\left<C_1,B_0,\mathcal{N}_2\right>$ and $\left<C_1,B_0,\mathcal{N}_4\right>$, respectively. The size of the architecture space is  $\approx 2.3298 \times 10^{13}$.

For the large search space, we have nine cells. The cell $C_0$ is the same as the one we used for the small search space. Each cell $C_i$, for $i \in [1, 8]$, has two blocks. The block $\left<C_i,B_0\right>$ has one \texttt{MLP\_Node}. 
The block $\left<C_i,B_1\right>$ has one \texttt{VariableNode} with the following set of \texttt{Connect} operations to create skip connections: Null, all combinations of inputs (i.e., 15 possibilities), all outputs of previous cells, and all $\mathcal{N}_0$ of previous cells except $C_0$.
Each $C_i$ takes as input the output of the cell $C_{i-1}$ for $i \in [1,\ldots, 8]$. The size of the architecture space is $\approx 5.7408 \times 10^{29}$. %

\subsubsection{\nt}

We define five types of nodes: \texttt{Conv\_Node}, \texttt{Act\_Node}, \texttt{Pool\_Node}, \texttt{Dense\_Node}, and \texttt{Dropout\_Node}. The \texttt{Conv\_Node} has the following options, where $x$ in Conv1D($x$) is the filter size. Here, the number of filters and the stride are set to 8 and 1, respectively: Identity, Conv1D(3), Conv1D(4), Conv1D(5), Conv1D(6).
The \texttt{Act\_Node} has the following options, where $x$ in Activation($x$) is a specific type of activation function: Identity, Activation($relu$), Activation($tanh$), Activation(($sigmoid$).
The \texttt{Pool\_Node} has the following options, where $x$ in MaxPooling1D($x$) represents the pooling size: Identity, MaxPooling1D(3), MaxPooling1D(4), MaxPooling1D(5), MaxPooling1D(6).
The \texttt{Dense\_Node} has the following options: Identity, Dense(10), Dense(50), Dense(100), Dense(200), Dense(250), Dense(500), Dense(750), Dense(1000).

The \texttt{Drop\_Node} has the following options: Identity, Dropout(0.5), Dropout(0.4), Dropout(0.3), Dropout(0.2), Dropout(0.1),\\ Dropout(0.05).

For the small search space, we define four cells: $C_0, C_1, C_2$, and $C_3$. Each cell $C_i$ has one block $\left<C_i,B_0\right>$, which takes the output of the previous cell as input except for the first block $\left<C_0,B_0\right>$, which  takes \texttt{RNA-seq gene expression} as input. This is followed by \texttt{CONV\_Node}, \texttt{ACT\_Nodes}, and \texttt{POOL\_Node}, which are connected sequentially. The blocks $\left<C_0,B_0\right>$ and $\left<C_1,B_0\right>$ have three sequentially connected \texttt{VariableNodes}: \texttt{Conv\_Node}, \texttt{Act\_Nodes}, and \texttt{Pool\_Node}. The blocks $\left<C_2,B_0\right>$ and $\left<C_3,B_0\right>$ have three sequentially connected \texttt{VariableNodes}: \texttt{Dense\_Node}, \texttt{Act\_Node}, and \texttt{Drop\_Node}. The size of the architecture space is $6.3504 \times  10^8$.

\subsection{Search strategy}

Different approaches have been developed to explore the space of neural architectures described by graphs. These approaches include random search, Bayesian optimization, evolutionary methods, RL, and other gradient-based methods. 
We focus on RL-based NAS, where an agent generates a neural architecture, trains the generated neural architecture on training data, and computes an accuracy metric on  validation data. The agent receives a positive  (negative) reward when the validation accuracy of the generated architecture increases (decreases). The goal of the agent is to learn to generate neural architectures that result in high validation accuracy by maximizing the agent's reward.

Policy gradient methods have emerged as a promising optimization approach for leveraging DL for RL problems \cite{sutton2018reinforcement,sutton2000policy}. These methods alternate between sampling and optimization using a loss function of the form
\begin{equation}
    J_t(\theta) =  \hat{\mathbb{E}}_t\lbrack \textrm{log}\pi_{\theta}(a_t|s_t)\hat{A}_t \rbrack,
\end{equation}
where $\pi_{\theta}(a_t|s_t)$ is a stochastic policy given by the action probabilities of a neural network (parameterized by $\theta$) that, for given a state $s_t$, performs an action $a_t$; $\hat{A}_t$ is the advantage function at time step $t$ that measures goodness of the sampled actions from $\pi_{\theta}$; and $\hat{\mathbb{E}}_t$ denotes the empirical average over a finite batch of sampled actions. The gradient of $J_t(\theta)$ is used in a gradient ascent scheme to update the neural network parameters $\theta$ to generate actions with high rewards. 

Actor-critic methods \cite{sutton2018reinforcement,grondman2012survey} improve the stability and convergence of policy gradient methods by using a separate critic to estimate the value of each state that serves as a state-dependent baseline. The critic is typically a neural network that progressively learns to predict the estimate of the reward given the current state $s_t$. The difference between the rewards collected at the current state from the policy network $\pi_{\theta}(a_t|s_t)$ and the estimate of the reward from the critic is used to compute the advantage. When the reward of the policy network 
is better (worse) than the estimate of a critic, the advantage function will be positive (negative), and the  
policy network parameters $\theta$
will be updated by using the gradient and the advantage function value.

Proximal policy optimization (PPO) is a policy gradient method for RL \cite{schulman2017proximal} that uses a loss function of the form 
\begin{equation}
    J_t(\theta) =  \hat{\mathbb{E}}_t\lbrack \textrm{min}(r_t(\theta)\hat{A}_t, \textrm{clip}(r_t(\theta), 1-\epsilon, 1+\epsilon) \hat{A}_t \rbrack,
\end{equation}
where $r_t(\theta) = \frac{\pi_\theta(a_t|s_t)}{\pi_{\theta_{\rm old}}(a_t|s_t)}$ is the ratio of action probabilities under the new and old policies; the clip/median operator ensures that the ratio is in the interval $[1-\epsilon, 1+\epsilon]$; and $\epsilon \in (0,1)$ is a hyperparameter (typically set to 0.1 or 0.2). The clipping operation prevents the sample-based stochastic gradient estimator of $\nabla_\theta J_t(\theta)$ from making extreme updates to
$\theta$.

\begin{figure}[t]
    \centering
    \includegraphics[width=0.45\textwidth]{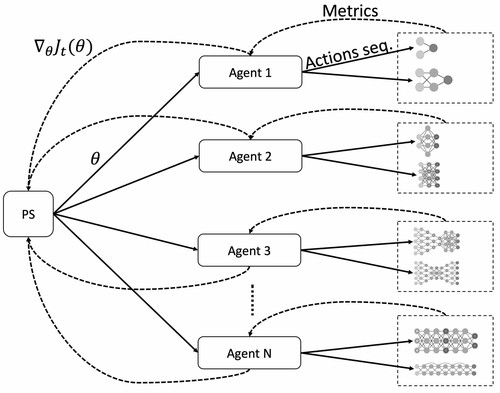}
    \caption{Synchronous and asynchronous manager-worker configuration for scaling 
    the RL-based NAS}
    \label{fig:manager-worker}
\end{figure}

We used two algorithmic approaches to scale up the NAS: synchronous advantage actor-critic (A2C) and asynchronous advantage actor-critic  (A3C). Both approaches use a manager-worker distributed learning paradigm as shown in Fig.~\ref{fig:manager-worker}. In A2C, all $N$ agents  start with the same policy network. 
At each step $t$, agent $i$ generates $M$ neural architectures, evaluates them in parallel (training and validation), and computes the gradient 
estimate 
$\nabla_\theta J_t(\theta)$ using the PPO method. Once the parameter server (PS) receives the PPO gradients from the $N$ agents, it averages the gradients and sends the result to each agent. The parameters of the policy network for each agent are updated by using the averaged gradient. The A3C method is similar to the A2C method except that an agent $i$ sends the PPO gradients to the PS, which does not wait for the gradients from all the agents before computing the average. Instead, the PS  computes the average from a set of recently received gradients and sends it to the agent $i$.

The synchronous update of A2C guarantees that the gradient updates to the PS are coordinated. 
A drawback of this approach is that the agents must wait until all $M * N$  tasks have completed in a given iteration.  Given a wide range of training times for the generated networks, A2C will underutilize nodes and limit parallel scalability.
On the other hand, A3C increases the node utilization at the expense of gradient staleness due to the asynchronous gradient updates, 
a staleness that grows
with the number of agents. 
While several works address synchronous and asynchronous updates in large batch supervised learning, studies in RL settings are limited, and none exists for the RL-based NAS methods studied here.

\subsection{Reward estimation strategy}
Crucial to the effectiveness of RL-based NAS is the way in which rewards are estimated for the agent-generated architectures. A naive approach of training each architecture from scratch on the full training data is computationally expensive even at scale and can require thousands of single-GPU days \cite{elsken2018neural}. A common approach to overcome this challenge is low-fidelity training, where the rewards are estimated by using a smaller number of training epochs \cite{zela2018towards}, a subset of original data \cite{klein2016learning}, a smaller proxy network for the original network \cite{zoph2018learning}, and a smaller proxy data set \cite{chrabaszcz2017downsampled}. In this paper, we use a smaller number of training epochs, a subset of the full training data, and timeout strategies to reduce the training time required for estimating the reward for architectures generated by NAS agents.

For the image data sets, research has showed that low-fidelity training can introduce a bias in reward estimation, which requires a gradual increase in fidelity as the search progresses \cite{li2016hyperband}. Whether this is the case for nonimage and nontext cancer data is not clear, however.
Moreover, the impact of low-fidelity training on NAS at scale 
is not well understood.
As we show next, at scale the RL-agent behavior (and consequently the generated architectures) exhibits different characteristics based on the fidelity level employed.

\section{Software Description}

\begin{figure}[t]
\centering
\includegraphics[width=0.45\textwidth]{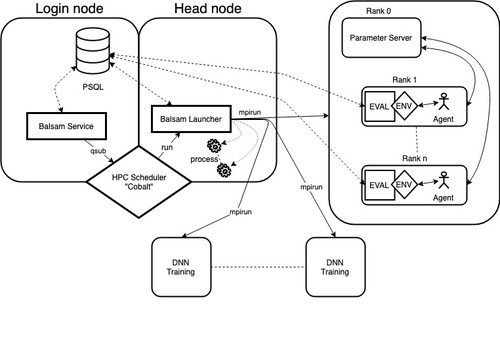}
\caption{
Distributed NAS architecture. The
Balsam service runs on a designated node, providing a
Django interface to a PostgreSQL database and interfacing
with the local batch scheduler for automated job submission. 
The launcher is a pilot job that runs on the allocated
resources and launches tasks from the database. The multiagent 
search runs as a single MPI application, and each agent
submits model evaluation tasks through the Balsam service
API. As these evaluations are added, the launcher continually executes them by
dispatch onto idle worker nodes. 
}
\label{fig:software-architecture}
\end{figure}
Our open source software comprises three Python subpackages: 
\texttt{benchmark}, a collection of representative NAS test problems;
\texttt{\mbox{evaluator}}, a model evaluation interface with several execution backends; 
and \texttt{search}, a suite of parallel NAS implementations that are implemented as 
distributed-memory \texttt{mpi4py} applications, where each MPI rank represents an RL agent. 

As new architectures are generated by the RL agents, the corresponding reward estimation tasks 
are submitted via the \texttt{evaluator} interface.  The \texttt{evaluator} exposes
a three-function API, 
that generically supports parallel asynchronous search methods. 
In the context of NAS, \texttt{add\_eval\_batch} submits new reward estimation tasks, while \texttt{get\_finished\_evals} is a nonblocking call that fetches newly completed reward estimations. 
This API enforces a complete separation
of concerns between the search and the backend for parallel evaluation of generated architectures. 
Moreover, a variety of \texttt{evaluator} backends, ranging from lightweight
threads to massively parallel jobs using a workflow system, allow a single search code to scale from 
toy models on a laptop to large DNNs running across leadership-class HPC resources.

We used the DeepHyper \cite{deephyper} software module on Theta, our target HPC platform, to dispatch reward estimation tasks to Balsam \cite{balsam}, a workflow manager enabling high-throughput, asynchronous task launching and monitoring for supercomputing platforms. Each agent exploited DeepHyper's evaluation cache and leveraged this to avoid repeating reward estimation tasks. A global cache of evaluated architectures is not maintained because that would nullify the benefit of agent-specific random weight initialization.  Balsam's performance monitoring capabilities are used to infer utilization as the fraction of allocated compute nodes actively running evaluation tasks at any given time $t$; the maximum value of 1.0 indicates that all worker nodes are busy evaluating configurations.

A schematic view of the NAS-Balsam infrastructure for parallel NAS is shown in Fig.~\ref{fig:software-architecture}. 
The \texttt{BalsamEvaluator} queries a Balsam PostgreSQL database through the Django model API.  Each NAS agent interacts with an environment that contains a \texttt{BalsamEvaluator};  therefore each agent has a separate database connection.  The Balsam launcher, in turn, 
pulls new reward estimation tasks and launches them onto idle nodes using a pilot-job mechanism. The launcher monitors ongoing tasks for completion
status and signals successful evaluations back to the \texttt{BalsamEvaluator}.

For the implementations of \async and \sync, we interfaced our NAS software with OpenAI Baselines \cite{baselines},  open-source software with a set of high-performing state-of-the-art RL methods. We developed an API for the RL methods in OpenAI Baselines so that we can leverage any new updates and RL methods that become available in the package. We followed the same interface as in OpenAI Gym \cite{openaigym} to create a NAS environment that encapsulates the \texttt{Evaluator} interface of Balsam to submit jobs for reward estimation.

The interface to specify the graph search space comprises support for structure, cell, block, variable node, constant node, mirror node, and operation. These are implemented as Python objects that allow the search space module to be extensible for different applications. The different choices for a given variable node are specified by using the \texttt{add\_op} method. These choices can be any set of Dense or Connect operations; the former creates a Keras \cite{chollet2017keras} dense layer and the latter creates skip connections. After a neural architecture is generated, the corresponding Keras model is created automatically for training and inference.

The analytics module of the software can be used to analyze the data obtained from  NAS. This module parses the logs from the NAS to extract the reward trajectory over time and to find the best architectures, worker utilization from the Balsam database, and number of unique architectures evaluated.

\section{Experimental results}

\begin{figure*}[t]
\begin{subfigure}{.33\textwidth}
  \centering
  \includegraphics[width=1.0\textwidth]{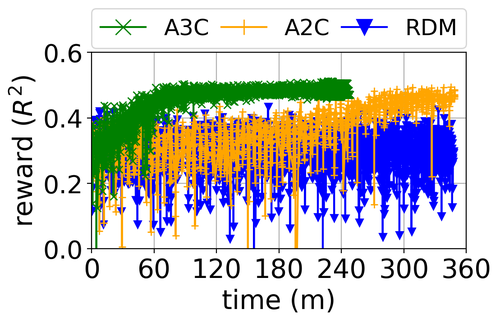}
  \caption{\combo}
  \label{fig:small-combo}
\end{subfigure}%
\begin{subfigure}{.33\textwidth}
  \centering
  \includegraphics[width=1.0\textwidth]{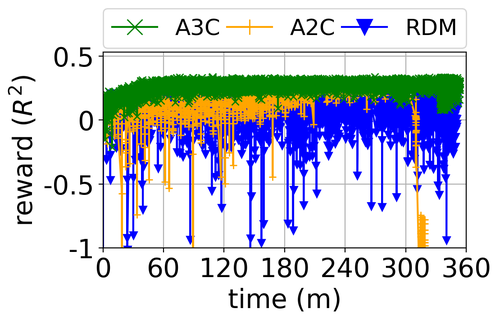}
  \caption{\uno}
  \label{fig:small-uno}
\end{subfigure}%
\begin{subfigure}{.33\textwidth}
  \centering
  \includegraphics[width=1.0\textwidth]{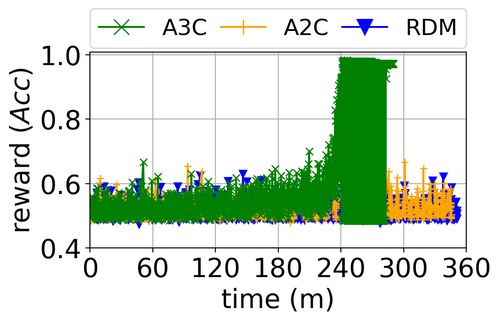}
  \caption{\nt}
  \label{fig:small-nt3}
\end{subfigure}%
\caption{Search trajectory showing reward over time for \async, \sync, and \rdm on the small search space}
\label{fig:search-small}
\vspace{-0.3cm}   
\end{figure*}

For the NAS search, we used \textrm{Theta}, a 4,392-node, 11.69-petaflop Cray XC40--based supercomputer at the Argonne Leadership Computing Facility (ALCF). Each node of Theta is a 64-core Intel Knights Landing (KNL) processor with 16 GB of in-package memory, 192 GB of DDR4 memory, and a 128 GB SSD. The compute nodes are interconnected by using an Aries fabric with a file system capacity of 10 PB. 

The reward estimation for a given architecture uses only a single KNL node (no distributed learning) with the number of training epochs set to 1 and timeout set to 10 minutes. 

The reward estimation for a generated architecture comprises two stages: training and validation. For \combo, the training is performed by using only 10\% of the training data. For \uno and \nt, since the data sizes are smaller, the full training data are used. The reward is computed by evaluating the trained model on the validation data set. For \combo and \uno, we use $R^2$ value as the reward; for $\nt$, we use classification accuracy (ACC). While we focus on accuracy in this paper,  other metrics can be specified, such as model size, training time, and inference time for a fixed accuracy using a custom reward function. 
To increase the exploration among agents, we used random weight initialization in the DNN training using agent-specific seeds during the reward estimation. Consequently, different agents generating the same architecture can have different rewards. For the policy network and value networks, we used a single-layer LSTM with 32 units and trained them with epochs=4, clip=0.2, and learning\_rate=0.001, respectively.

Once the NAS search was completed on \textrm{Theta}, we selected the top $50$ DNN architectures from the  search based on the estimated reward values. We performed post-training, where we trained the DNNs for a larger number of epochs (20 for all experiments), without a timeout, and on the full training data. 
Running post-training on the KNL nodes was slower; therefore we used \textrm{Cooley}, a GPU cluster at the ALCF. \textrm{Cooley} has 126 compute nodes; each node has 12 CPU cores, one NVIDIA Tesla K80 dual-GPU card, with 24 GB of GPU memory and 384 GB of DDR3 CPU memory. 
The manually designed \combo network took 2215.13 seconds and  705.26 seconds for training on KNL and K80 GPUs, respectively. We ran 50 models on 25 GPUs with two model trainings per K80 dual-GPU card, one per GPU. For both the reward estimation and post-training, we used the Adam optimizer with a default learning rate of 0.001. The batch size was set to 256, 32, and 20 for \combo, \uno, and \nt, respectively. The same values were used in the manually designed networks. %

We evaluated the generated architectures after post-training with respect to three metrics:  
 {\bf accuracy ratio ($R^2/R^2_b$ or $ACC/ACC_b$)}, given by the ratio of $R^2$ ($ACC$) of a NAS-generated architecture and the manually designed  network for \combo and \uno (\nt);
{\bf trainable parameters ratio ($P_b/P$)}, given by the ratio of number of trainable parameters of the manually designed network and the given NAS-generated architecture (this metric helped us  evaluate the ability of  NAS to build smaller networks, which have  better generalization and fewer overfitting issues compared with larger networks; and
{\bf training time ratio ($T_b/T$)}, given by the ratio of the post-training time (on a single NVIDIA Tesla K80 GPU) of the manually designed network and the given NAS-generated architecture  
(this metric allows us to evaluate the ability of the NAS to find faster-to-train networks, which are useful for hyperparameter search and subsequent training with additional data).

The Theta environment consists of Cray Python 3.6.1, TensorFlow 1.13.1 \cite{abadi2016tensorflow}. Based on ALCF recommendations, we used the following environment variable settings to increase the performance of TensorFlow: \texttt{KMP\_BLOCKTIME='0'}, \\ 
\texttt{KMP\_AFFINITY='granularity=fine,compact,1,0'}, \\
and \texttt{intra\_op\_parallelism\_threads=62}. The Cooley environment consists of Intel Python 3.6.5, Tensorflow-GPU 1.13.1.

\subsection{Evaluation of the search strategy}

In this section, we show that in spite of gradient staleness issue, \async has a faster learning capability and a better system utilization than does \sync; synchronized gradient updates and the consequent node idleness adversely affect the efficacy of \sync. 

We evaluated the learning and convergence capabilities of \async and \sync by comparing them with random search (\rdm), where agents perform actions at random and will not compute and synchronize gradients with the parameter server. This comparison was to ensure that the search space was the same as \async, \sync, and \rdm and  allowed us to evaluate the search capabilities of \async and \sync with all other settings remaining constant. We used 256 Theta nodes for \async, \sync, and \rdm with 21 agents and 11 workers per agent\footnote{We want to set number of agents $\approx 2$ $\times$ number of workers: 21 agents and 11 workers satisfy the constraint with minimal unused nodes. We requested 256 instead of 253 to get 6 hours of running time.}: 21 agent nodes, 231 worker nodes, 1 Balsam workflow node, and 3 unused nodes.

Figure \ref{fig:search-small} shows rewards obtained over time for \async, \sync, and \rdm. We observe that \async outperforms \sync and \rdm with respect to both time and rewards obtained. \async exhibits a faster learning trajectory than does  \sync and reaches a higher reward in a shorter wall-clock time. \async reaches reward values of $0.5$ and $0.4$ in approximately 70 and 35 minutes for \combo and \uno, respectively, after which the increase in the reward values is  small. On \combo and \nt, \async ends in 250 and 285 minutes, respectively, because all the agents generate the same architecture for which the agent-specific cache returns the same reward value. We detected this and stopped the search since it could not proceed in a meaningful way. On \uno, \async generates different architectures and  does not end before the wall-clock time limit. \sync shows a slower  learning trajectory; it eventually reaches the reward values of \async on \combo and \uno, but its reward value on \nt is  poor. As expected, \rdm shows neither learning capability nor the ability to reach higher reward values. On \nt, we found an oscillatory behavior with \async toward the end. After finding higher rewards, \async did not converge as expected. After a closer examination of the architectures and their reward values, we found that although the agents are producing similar architectures, the reward estimation is  sensitive to a random initializer with one training epoch and a batch size of 20. Consequently, the same architecture produced by two different agents had significantly different rewards (e.g., 1.0 and 0.4). 

\begin{figure*}[t]
\begin{subfigure}{.33\textwidth}
  \centering
  \includegraphics[width=1.0\textwidth]{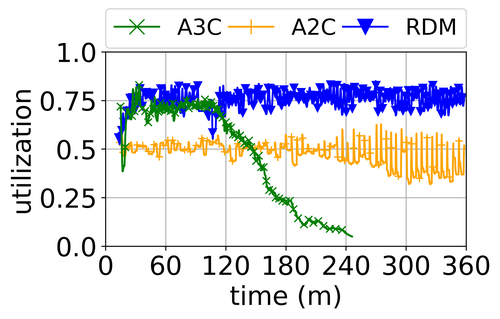}
  \caption{\combo}
  \label{fig:small-combo2}
\end{subfigure}%
\begin{subfigure}{.33\textwidth}
  \centering
  \includegraphics[width=1.0\textwidth]{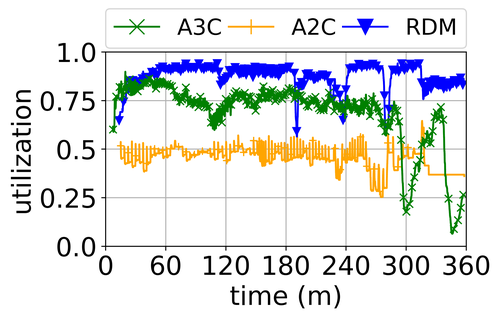}
  \caption{\uno}
  \label{fig:small-uno2}
\end{subfigure}%
\begin{subfigure}{.33\textwidth}
  \centering
  \includegraphics[width=1.0\textwidth]{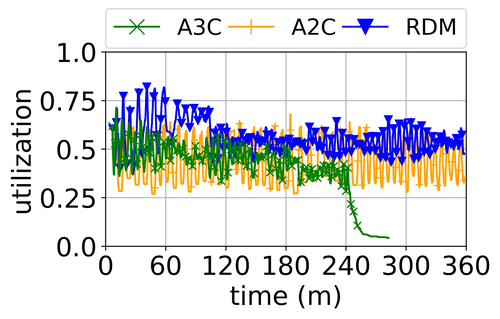}
  \caption{\nt}
  \label{fig:small-nt32}
\end{subfigure}%
\caption{Utilization for \async, \sync, and \rdm on the small search space}
\label{fig:node-util-small}
\vspace{-0.3cm}   
\end{figure*}
Figure \ref{fig:node-util-small} shows the utilization over time for \async, \sync, and \rdm on the small search space. 
The utilization of the \rdm on \combo stays at $1.0$ in the initial search stages, but after that it averages $0.75$.  Although \rdm lends itself to an entirely asynchronous search, the estimation of $M$ rewards per agent was blocking in our implementation.  This per-agent synchronization, combined with variability of the reward estimation times, leads to suboptimal utilization.  
The utilization of \async is similar to that of \rdm until 100 minutes, after which there is a steady decrease due to an increase in the caching effect; this is just a manifestation of the convergence of \async, which stops after 160 minutes.

On \uno, the utilization of \rdm becomes high, with an average of $0.9$. This is because randomly sampled DNNs in this space have a smaller variance of reward estimation times. The utilization of \async is similar to that of \rdm in the beginning of the search, but it decreases after 50 minutes because it learns to generate architectures that have a shorter training time with higher rewards. On \nt, the utilizations of \rdm and \sync are similar to that of \combo but with even lower values because per batch  several architectures have a shorter reward estimation time. The utilization of \sync shows a sawtooth shape; because of the synchronous nature, at the start of each batch the utilization goes to 1, then drops off and becomes zero when all agents finish their batch evaluation. 
\begin{figure}[h]
\begin{subfigure}{.25\textwidth}
  \centering
  \includegraphics[width=1.0\textwidth]{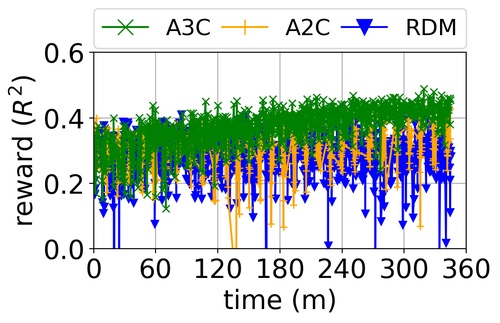}
  \caption{Search trajectory}
  \label{fig:large-combo-search}
\end{subfigure}%
\begin{subfigure}{.25\textwidth}
  \centering
  \includegraphics[width=1.0\textwidth]{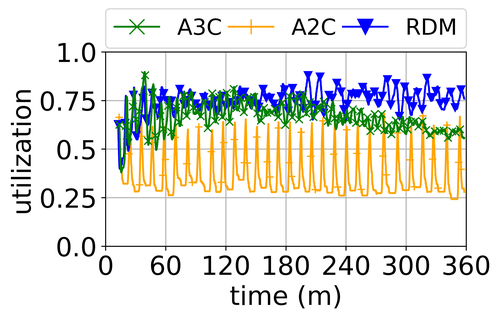}
  \caption{Utilization}
  \label{fig:large-combo-util}
\end{subfigure}%
\caption{Results on \combo with the large search space}
\label{fig:combo-large}
\vspace{-0.6cm}   
\end{figure}

Figure \ref{fig:combo-large} shows the search trajectory and utilization of \async on \combo with the large search space. We observe that \async finds higher rewards faster than do \sync and \rdm. The utilization of \async is similar to that of \rdm (75\% average) until 200 minutes, after which there is a gradual decrease because of the caching effect. Nevertheless, the search did not converge and stop as it did in the small search space.

\subsection{Comparison of \async-generated networks with manually designed  networks}

\begin{figure}
\begin{subfigure}{.23\textwidth}
  \centering
  \includegraphics[width=1.0\textwidth]{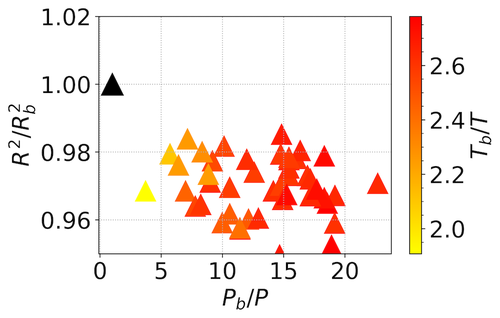}
  \caption{\combo}
  \label{fig:small1}
\end{subfigure}%
\begin{subfigure}{.23\textwidth}
  \centering
  \includegraphics[width=1.0\textwidth]{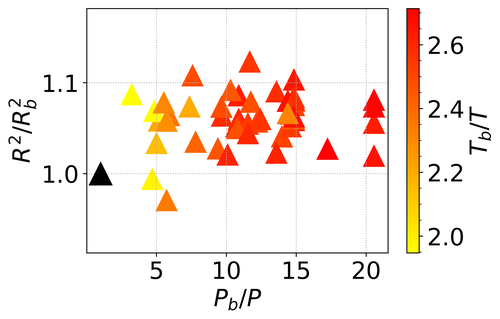}
  \caption{\uno}
  \label{fig:small2}
\end{subfigure}\\
\begin{subfigure}{.23\textwidth}
  \centering
  \includegraphics[width=1.0\textwidth]{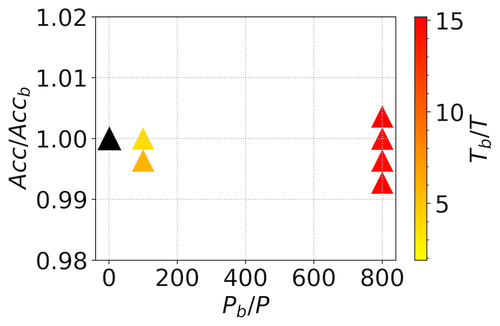}
  \caption{\nt}
  \label{fig:small3}
\end{subfigure}
\caption{
Post-training results on the top 50 \async architectures from the small search space run on 256 nodes. 
Accuracy ratios ($R^2/R^2_b$, $Acc/Acc_b$) $>1.0$ indicate  a A3C-generated architecture outperforming the manually designed network. Trainable parameter ratios ($P_b/P$) $ > 1.0$ indicate that a A3C-generated architecture has fewer trainable parameters than the manually designed network. Training time ratios ($T_b/T$) $> 1.0$ indicate that a A3C-generated architecture is faster than the manually designed network.
}
\label{fig:post-train-small}
\vspace{-0.5cm}
\end{figure}

\begin{figure}
\begin{subfigure}{.25\textwidth}
  \centering
   \includegraphics[width=1.0\textwidth]{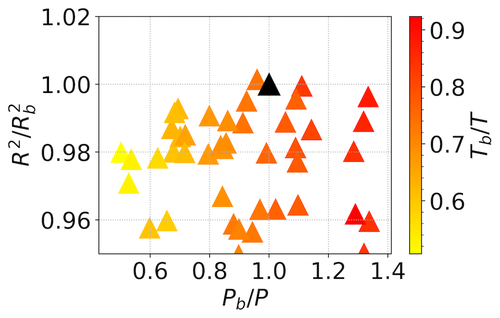}
  \caption{\combo}
  \label{fig:large1}
\end{subfigure}%
\begin{subfigure}{.25\textwidth}
  \centering
  \includegraphics[width=1.0\textwidth]{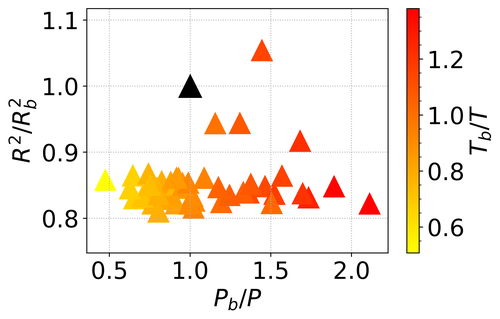}
  \caption{\uno}
  \label{fig:large2}
\end{subfigure}\\

\caption{Post-training results on top-50 \async architectures from the large search space run on 256 nodes}
\label{fig:post-train-large}
\vspace{-0.5cm}
\end{figure}
Here, we show that \async discovers architectures that have significantly fewer trainable parameters, shorter training time, and accuracy similar to or higher than those of manually designed architectures.

Figure~\ref{fig:post-train-small} shows the post-training results of \async on the 50 best architectures selected based on the estimated reward during the NAS. 
From the accuracy perspective, on \combo, five \async-generated architectures obtain $R^2$ values that are competitive with the manually designed network ($R^2/R^2_b > 0.98$); on \uno, more than forty \async-generated architectures obtain $R^2$ values that are better than the manually designed network value ($R^2/R^2_b > 1.0$); on \nt, three \async-generated architectures obtain accuracy values that are higher than that of the manually designed network ($ACC/ACC_b > 1.0$). From the trainable parameter ratio viewpoint, \async-generated architectures completely outperform the manually designed networks on all three data sets. On \combo, \async-generated architectures have 5x to 15x fewer trainable parameters than  the manually designed network has; on \uno this is between 2x to 20x; on \nt, \async-generated architectures have up to 800x fewer parameters than  the manually designed network has. The significant reduction in the number of trainable parameters is reflected in the training time ratio, where we observed up to 2.5x speedup for \combo and \uno and up to 20x for \nt.

Figure~\ref{fig:post-train-large} shows the post-training results of \async with the large search space on \combo and \uno. 
On \combo,  use of the large search space allowed \async to generate a number of architectures with accuracy values  higher than those generated with the small search space. Among them, five architectures obtained $R^2/R^2_b > 0.99$; one  was better than the manually designed network. 
The large search space increases the number of training parameters and training time significantly. Nevertheless, on \uno, we found that the larger search space decreases the accuracy values significantly, which can be attributed to the overparameterization given the relatively small amount of data, and additional improvement in accuracy was not observed after a certain number of epochs.

\subsection{Scaling \async on \combo with large search space}
In this section, we demonstrate that increasing the number of agents and keeping the number of workers to a smaller value in \async result in better scalability and improvement in accuracy. 

We ran \async on \combo with the large search space on $512$ and $1,024$ KNL nodes.\footnote{We did not use more than 1,024 nodes in this experiment because of a system policy limiting the total number of concurrent application launches to 1,000.}  We studied two approaches to scaling. In the first approach, called \emph{worker scaling},  we fixed the number of agents at 21 and varied the number of workers per agent.  For 512 and 1,024 nodes, we tested 23 and 47 workers per agent, respectively.  In the second approach, called \emph{agent scaling},  we fixed the number of workers per agent at 11 and increased the number of agents. For 512 and 1,024 nodes, we used 42 and 85 agents, respectively. %

\begin{figure}[t]
    \centering
    \includegraphics[width=0.5\textwidth]{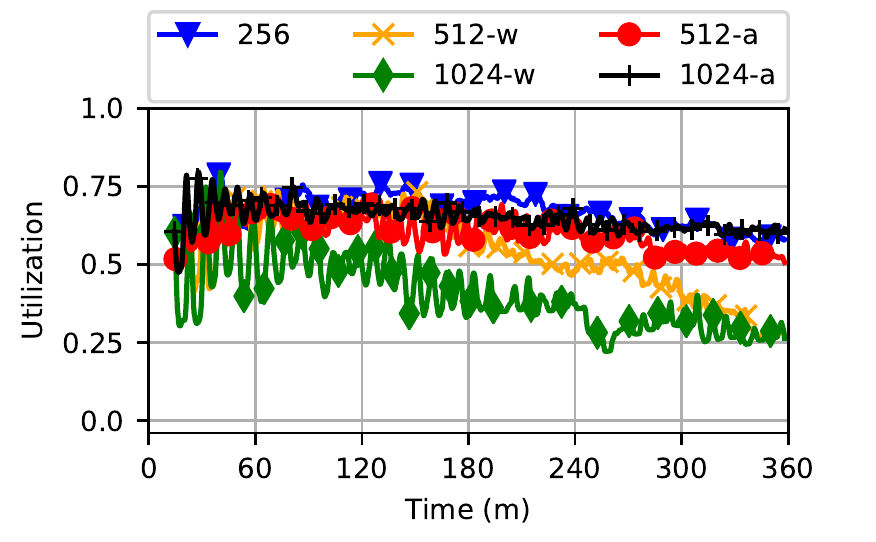}
    \caption{
    Utilization of \async on \combo with the large search space run on 512 and 1,024 nodes with agent and worker scaling; 256 nodes are used as reference.
    }
    \label{fig:combo-scaling}
\vspace{-0.5cm}
\end{figure}
The utilization of \async is shown in Fig.~\ref{fig:combo-scaling}. We observe that scaling the number of agents is more efficient than is scaling the number of workers per agent. In particular, the utilization values of agent scaling, 512(a) and 1,024(a), are similar to those measured at 256 nodes; there is no significant loss in utilization by going to higher node counts.  On the other hand, utilization suffers as the number of workers per agent is increased. The reason is that the worker evaluations are batch synchronous and the increase in workers results in an increase in the number of idle nodes within a batch. The decreasing overall trend in utilization can be attributed to the increased cache effect.

\begin{figure}
\begin{subfigure}{.23\textwidth}
  \centering
  \includegraphics[width=1.0\textwidth]{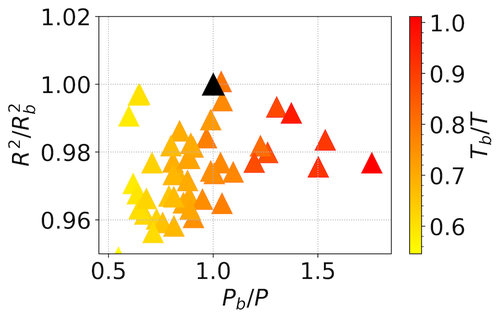}
  \caption{512 nodes}
  \label{fig:sfig3}
\end{subfigure}
\begin{subfigure}{.23\textwidth}
  \centering
  \includegraphics[width=1.0\textwidth]{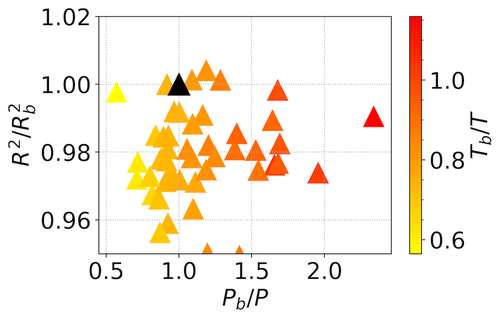}
  \caption{1,024 nodes}
  \label{fig:sfig4}
\end{subfigure}
\caption{Post-training results of A3C on \combo with the large search space  run on 512 and 1,024 nodes with agent scaling}
\label{fig:scaling}
\vspace{-0.6cm}
\end{figure}
Figure \ref{fig:scaling} shows the post-training results of the 50 best architectures from the 51-2 and 1,024-node agent scaling experiments. Compared with the 256-node experimental results (see Fig.~\ref{fig:large1}), both the 512-node and 1,024-node experiments result in network architectures that have better accuracy, fewer trainable parameters, and lower training time. In particular, scaling on 1,024 nodes results in nine networks with $R^2/R^2_b > 0.99$; among them four  networks were better than the manually designed network. These networks have as few as 50\% fewer parameters than the manually designed network has. An increase in the number of nodes and agents results in higher exploration of the architecture space, which eventually increases the chances of finding a diverse range of architectures without sacrificing accuracy.

\subsection{Impact of fidelity in reward estimation}

Here, we show that, at scale, the fidelity of the reward estimation affects agent learning behavior in different ways and can generate diverse architectures. 

We analyzed the impact of fidelity in the reward estimation strategy by increasing the training data size in \async from the default of 10\% to 20\%, 30\%, and 40\% on \combo. We ran the experiments on 256 nodes and used the default values for the training epochs and the timeout. 

\begin{figure}[h]
    \centering
    \includegraphics[width=0.5\textwidth]{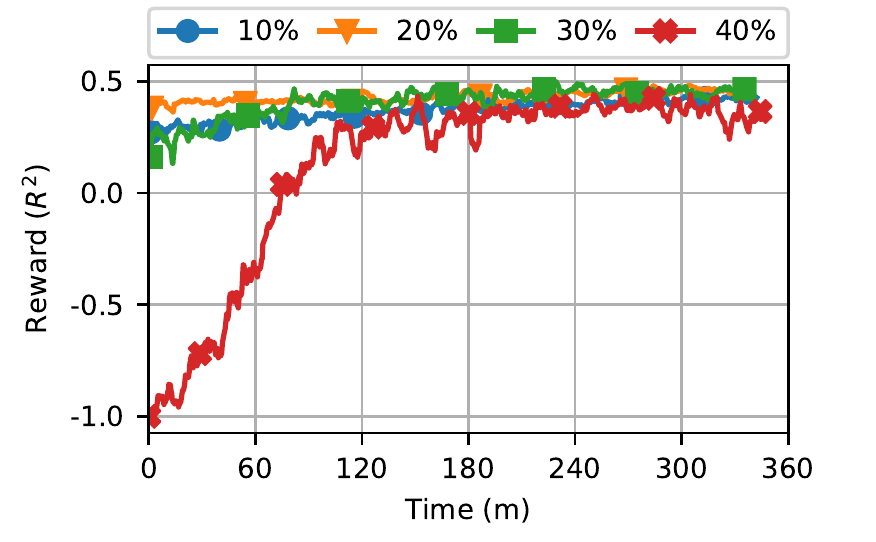}
    \caption{Rewards over time obtained by \async for \combo with the large search space on 256 nodes with different training data sizes}
    \label{fig:tdata-search}
\vspace{-0.5cm}
\end{figure}

Figure \ref{fig:tdata-search} shows the search trajectory of \async. We can observe that on 10\%, 20\%, and 30\% training data, \async generates architectures with high rewards within 80 minutes. With 40\% training data, the improvement in the reward is  slow. The reason  is that the large number of architectures generated by \async cannot complete training before the timeout. Consequently, it takes 80 minutes to reach reward values greater than 0. Nevertheless, it slowly learns to generate architectures that can be trained within the timeout---within 160 minutes, \async 
with 40\% of the training data reaches the reward values found by \async with less training data.
 
\begin{figure}
\begin{subfigure}{.23\textwidth}
  \centering
    \includegraphics[width=1.0\textwidth]{results-img/large/hex_training_pt_256.png}
  \caption{10\%}
  \label{fig:tdata1}
\end{subfigure}%
\begin{subfigure}{.23\textwidth}
  \centering
  \includegraphics[width=1.0\textwidth]{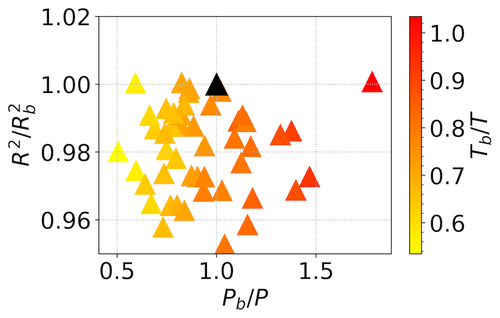}
  \caption{20\%}
  \label{fig:tdata2}
\end{subfigure}%
\\
\begin{subfigure}{.23\textwidth}
  \centering
  \includegraphics[width=1.0\textwidth]{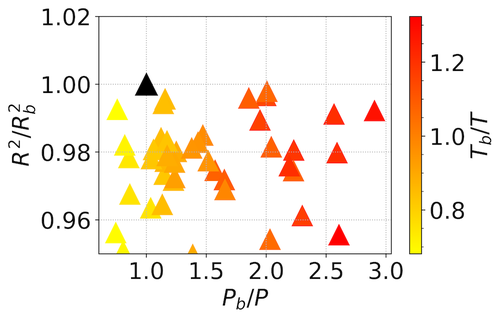}
  \caption{30\%}
  \label{fig:tdata3}
\end{subfigure}
\begin{subfigure}{.23\textwidth}
  \centering
  \includegraphics[width=1.0\textwidth]{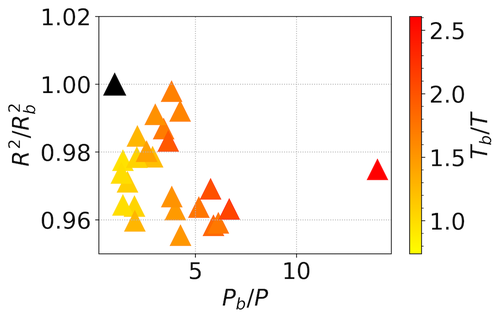}
  \caption{40\%}
  \label{fig:tdata4}
\end{subfigure} 
\caption{Post-training results of A3C on \combo with a large search space run on 256 nodes}
\label{fig:tdata}
\vspace{-0.6cm}
\end{figure}
The post-training results are shown in Fig.~\ref{fig:tdata}. As we increase the training data size in the reward estimation, we can observe a trend in which the best architectures generated by \async have fewer trainable parameters and shorter post-training time. In the 10\% case, the training time in reward estimation is not a bottleneck. Consequently, the agents generate networks with fewer trainable parameters to increase the reward, and the post-training time of the best architectures often exceeds that of the manually designed  network. 
Increasing the training data size to 20\% results in the best architectures that have smaller trainable parameters and longer post-training time than the manually designed network has. %
We found that the agents that can achieve faster rewards by using fewer parameters update the parameter server and bias the search. In the 30\% case, the training time affects the best architectures. Consequently, several best architectures have fewer trainable parameters and shorter post-training time than the 10\% and 20\% cases have. In the 40\% case,  the training time in the reward estimation becomes a bottleneck. As a result, the agent learns to maximize the reward by generating architectures with faster training time in the reward estimation by using fewer trainable parameters.

\subsection{Impact of randomness in \async}
The \async strategy that we used in  NAS is a randomized method. The randomness stems from several sources, including  random weight initialization of the neural networks, asynchronicity, and stochastic gradient descent for reward estimation. 
Here, we analyze the impact of randomness on the search trajectory of \async. We repeated \async 10 times on the \combo benchmark with the small search space.
\begin{figure}[t]
    \centering
    \includegraphics[width=0.5\textwidth]{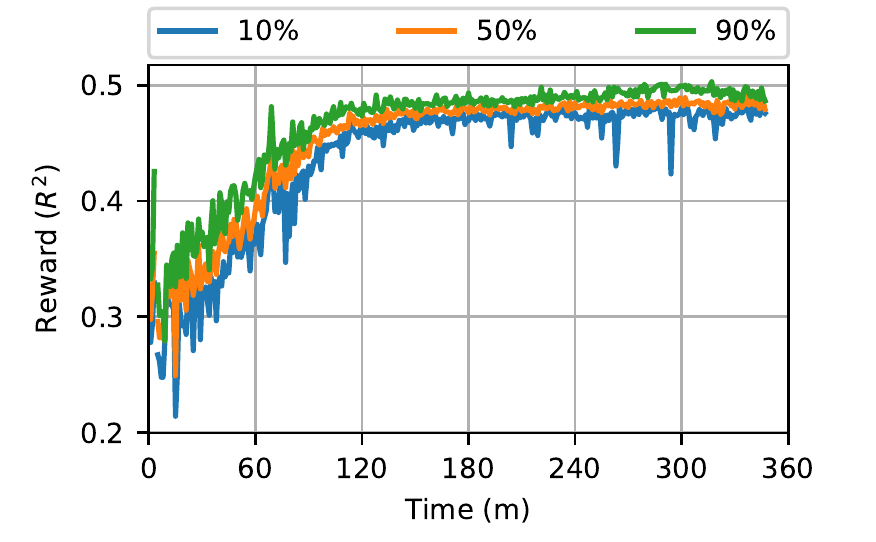}
    \caption{Statistics of the \async search trajectory computed over 10 replications on \combo with small search space}
    \label{fig:combo-replications}
\vspace{-0.3cm}    
\end{figure}
The results are shown in Fig.~\ref{fig:combo-replications}. Given a time stamp, we compute 10\%, 50\% (median), and 90\% quantiles from 10 values---this removes both the best and worst values (outliers) for a given time stamp. In the beginning of the search, the differences in the search trajectory of \async are noticeable, where the quantiles of the reward range between 0.2 and 0.4. Nevertheless, the variations become smaller as the search progresses. At the end of the search, all the quantile values are close to $0.5$, indicating that the search trajectories of different replications are similar and the randomness does not have a significant impact on the search trajectory of \async.

\subsection{Summary of the best A3C architectures}

\begin{table}
\caption{Summary of best architectures found by A3C} \label{tab:results-summary}
\vspace{-0.5cm}
\begin{tabular}{lrrrl}
                  & Trainable      & Training     & &  \\ 
                  & Parameters      & Time (s)    &  $R^2$ or $ACC$      &  \\ 
\cline{1-4}
\multicolumn{5}{c}{Combo} \\
\cline{1-4}
manually designed         & 13,772,001             & 705.26                  & 0.926  &\\ 
\textbf{A3C-best} & \textbf{1,883,301}     & \textbf{283.00}         & \textbf{0.93}   &\\ 
\cline{1-4}
\multicolumn{5}{c}{Uno}\\
\cline{1-4}
manually designed          & 19,274,001             & 164.94                  & 0.649                &  \\
\textbf{A3C-best} & \textbf{1,670,401}     & \textbf{63.53}          & \textbf{0.729}       &  \\ \cline{1-4}
\multicolumn{5}{c}{NT3} \\
\cline{1-4}
manually designed          & 96,777,878             & 247.63                  & 0.986                &  \\ 
\textbf{A3C-best} & \textbf{120,968}      & \textbf{16.65}          & \textbf{0.989}       &  \\ \cline{1-4}
\end{tabular}
\end{table}
Table \ref{tab:results-summary} summarizes the results of the best A3C-generated architectures with respect to the manually designed networks on three data sets.
On \combo, the accuracy of the best A3C-generated architecture is slightly better than that of the manually designed network. However, it has  
7.3x fewer trainable parameters and 2.5x faster training time.
O%
On \uno, the best A3C-generated architecture  outperforms the manually designed network with respect to all three factors:  trainable parameters, training time, and accuracy.  
The best NAS architecture obtained an $R^2$ value of $0.729$ with 11.5x fewer trainable parameters and 2.5x faster training time than that of the manually designed network.  On \nt, the best A3C-generated network obtained 98\% accuracy (similar to the manually designed network), but it has 800x fewer trainable parameters and 14.8x faster training time. 
Moreover, whereas the manual design of networks for these data sets took days to weeks, a NAS run with our scalable open-source software 
will take only six hours of wall-clock time for similar data sets.

\vspace{-0.1cm}
\section{Related work}
We refer the reader to \cite{nasonlinebib2019,elsken2018neural,distdl-preprint} for a detailed exposition on NAS related work. Here, to highlight our contributions,  we discuss related work across five dimensions. 

{\bf Application:} A majority of the NAS literature focuses on the automatic construction of CNNs for image classification tasks applied primarily to benchmark data sets such as CIFAR and ImageNet. This is followed by recurrent neural nets for text classification tasks on benchmark data sets. Application of new domain applications beyond standard benchmark tasks is still in its infancy \cite{elsken2018neural}. Recent examples include language modeling \cite{zoph2016neural}, music modeling \cite{rawal2018nodes}, image restoration \cite{suganuma2018exploiting}, and network compression \cite{ashok2017n2n} tasks. While there exist several prior works on DL for cancer data, we believe our work is the first application of NAS for cancer predictive modeling tasks.

{\bf Search space:} Two key elements define the search space: primitives and the architecture template. Existing works have used convolution with different numbers of filters, kernel size, and strides; pooling such as average and maximum depthwise separable convolutions; dilated convolutions; RNN' LSTM cells; and a layer with a number of units for fully connected networks. Motivated by the requirements of the cancer data, we introduced new types of primitives such as multiple input layers, variable nodes, fixed nodes, and mirror nodes, which allow us to explicitly incorporate cancer domain knowledge in the search space. Existing architecture templates range from simple chain-structured skip connections to cell-based architectures \cite{elsken2018neural}. The NAS search space that we designed is not specific to a single template, and it can handle all three template types. 
More important, we can define templates that enable a search over cells specific to the cancer data.

{\bf Search method:} Different search methods have been used to navigate the search space, such as random search \cite{bergstra2012random,li2019random,sciuto2019evaluating}, Bayesian optimization \cite{snoek2012practical,bergstra2013hyperopt,klein-bayesopt17,dikov2019bayesian,kandasamy2018neural,rohekar2018constructing,wang2018combination,wistubabayesian}, evolutionary methods 
\cite{chen2018reinforced,chu2019fast,liang2018evolutionary,liang2019evolutionary,lorenzo2018memetic,maziarz2018evolutionary,stanley2019,suganuma2018exploiting,van2018evolutionary,young2017evolving,patton2018167}, gradient-based methods, and reinforcement learning \cite{ashok2017n2n,baker2016designing,bello2017neural,bender2018understanding,chu2019fast,guo2018irlas,pmlr-v80-bender18a,stanley2019,xie2018snas,zoph2016neural}. Currently, there is no clear winner (for example, see \citet{real2018regularized});
most likely a single method may never outperform all other methods on all data sets under all possible settings (as a consequence of the "no free lunch" theorem). Therefore, we need to understand the strengths and limitations of these methods based on the data set. We compared A3C and A2C methods at scale and analyzed their  convergence on nontext and nonimage data sets. We showed that A3C, despite gradient staleness due to asynchronous gradient update, can find high-performing DNNs in a short computation time. We evaluated the efficacy of the RL-based NAS at scale with respect to accuracy, training time, parameters, and reward estimation fidelity.

{\bf RL-based NAS scalability:} In \cite{zoph2016neural}, RL-based NAS was scaled on 800 GPUs using 20 parameter servers, 100 agents, and 8 workers per agent. This approach was run for three to four weeks. In \cite{zoph2018learning}, a single RL agent generated 450 networks and used 450 GPUs for concurrent training across four days. In both these works, the primary goal was to demonstrate that the NAS can outperform manually designed networks on image and text classification tasks. We demonstrated RL-based NAS experiments on up to 1,024 KNL nodes and for a much shorter wall-clock time of 6 hours on cancer data.

{\bf Open source software:} AutoKeras \cite{jin2018efficient} is 
an open source automated machine learning  package that uses Bayesian optimization and  network morphism for NAS. The scalability of the package is limited because it is designed to run on single node with multiple GPUs that can evaluate few architectures in parallel. Microsoft's Neural Network Intelligence \cite{nni2019} is a open source AutoML package  designed primarily to tune the hyperparameters of a fixed DNN architecture by using different types of search methods; it lacks capabilities for architecture templates. 
Ray \cite{moritz2018ray} is a open source  high-performance distributed execution framework that has modules for RL and hyperparameter search but does not have support for NAS.  
AMLA \cite{kamathamla} is a framework for implementing and deploying AutoML neural network generation algorithms. Nevertheless, it has not been demonstrated  on benchmark applications or at scale. 
TPOT \cite{OlsonGECCO2016} optimizes scikit-learn \cite{scikit-learn}, a library of classical machine learning algorithms, using evolutionary algorithms, but it does not have support for NAS.   
Our package differs from the existing ones with respect to customized NAS search space for cancer data and  scalabiltiy. 
\vspace{-0.25cm}

\section{Conclusion and future work}

We developed scalable RL-based NAS to automate DNN model development for a class of cancer data. We designed a NAS search space that takes into account characteristics specific to nonimage and nontext cancer data. We scaled the proximal policy optimization, a state-of-the-art RL approach, using a manager-worker approach on up to 1,024 Intel Knights Landing nodes and evaluated its efficacy using cancer DL benchmarks. We demonstrated the efficacy of this method at scale and showed that the asynchronous actor critic method (\async) outperforms its synchronous and random variants. The results showed that \async can discover architectures that have significantly fewer training parameters, shorter training time, and accuracy similar to or higher than those of manually designed architectures. 
We experimented with the volume of training data in reward estimation, analyzed the impact of fidelity in reward estimation on the agent learning capabilities, and showed that it can be used to discover different types of network architectures. 

Our future work will include applying NAS on a broader class of cancer data, conducting an architecture search for transformer and attention-based networks \cite{vaswani2017attention,luong2015effective}, adapting NAS for  multiple objectives, developing adaptive reward estimation approaches, developing multiparameter servers to improve scalability, integrating hyperparameter search approaches, and comparing our approach with extremely scalable evolutionary approaches such as MENNDL \cite{young2017evolving,patton2018167} and Bayesian optimization methods \cite{snoek2015scalable}. Our NAS framework is designed to be flexible for developing surrogate DNN models for tabular data. We will explore NAS for reduced-order modeling in scientific application areas such as climate and fluid dynamics simulations. 

NAS has the potential to accelerate cancer deep learning research. A scalable open-source NAS package such as ours can allow cancer researchers to automate neural architecture discovery using HPC resources and to experiment with diverse DNN architectures. This will be of paramount importance in order to tackle cancer as we incorporate more diverse and complex data sets.

\section*{Acknowledgment}
This material is based upon work supported by the U.S.\ Department of Energy 
(DOE), Office of Science, Office of Advanced Scientific Computing Research, under
Contract DE-AC02-06CH11357. This research used resources of the Argonne 
Leadership Computing Facility, which is a DOE Office of Science User Facility.

\bibliographystyle{ACM-Reference-Format}
\bibliography{bibs/pbalapra-hps,bibs/nas}


\begin{thebibliography}{89}


\ifx \showCODEN    \undefined \def \showCODEN     #1{\unskip}     \fi
\ifx \showDOI      \undefined \def \showDOI       #1{#1}\fi
\ifx \showISBNx    \undefined \def \showISBNx     #1{\unskip}     \fi
\ifx \showISBNxiii \undefined \def \showISBNxiii  #1{\unskip}     \fi
\ifx \showISSN     \undefined \def \showISSN      #1{\unskip}     \fi
\ifx \showLCCN     \undefined \def \showLCCN      #1{\unskip}     \fi
\ifx \shownote     \undefined \def \shownote      #1{#1}          \fi
\ifx \showarticletitle \undefined \def \showarticletitle #1{#1}   \fi
\ifx \showURL      \undefined \def \showURL       {\relax}        \fi
\providecommand\bibfield[2]{#2}
\providecommand\bibinfo[2]{#2}
\providecommand\natexlab[1]{#1}
\providecommand\showeprint[2][]{arXiv:#2}

\bibitem[\protect\citeauthoryear{??}{aut}{[n. d.]a}]%
        {automl}
 \bibinfo{year}{[n. d.]}\natexlab{a}.
\newblock \bibinfo{title}{{AutoML} Workshops}.
\newblock
\newblock
\urldef\tempurl%
\url{https://www.ml4aad.org/automl/}
\showURL{%
\tempurl}


\bibitem[\protect\citeauthoryear{??}{can}{[n. d.]a}]%
        {candle-code}
 \bibinfo{year}{[n. d.]}\natexlab{a}.
\newblock \bibinfo{title}{{CANDLE Exascale Computing Program Application}}.
\newblock
\newblock
\urldef\tempurl%
\url{https://github.com/ECP-CANDLE/Benchmarks}
\showURL{%
\tempurl}


\bibitem[\protect\citeauthoryear{??}{com}{[n. d.]}]%
        {combo-code}
 \bibinfo{year}{[n. d.]}\natexlab{}.
\newblock \bibinfo{title}{{Combo Benchmark}}.
\newblock
\newblock
\urldef\tempurl%
\url{https://github.com/ECP-CANDLE/Benchmarks/tree/master/Pilot1/Combo}
\showURL{%
\tempurl}


\bibitem[\protect\citeauthoryear{??}{aut}{[n. d.]b}]%
        {automl-google}
 \bibinfo{year}{[n. d.]}\natexlab{b}.
\newblock \bibinfo{title}{An End-to-End AutoML Solution for Tabular Data at
  KaggleDays}.
\newblock
\newblock
\urldef\tempurl%
\url{https://ai.googleblog.com/2019/05/an-end-to-end-automl-solution-for.html}
\showURL{%
\tempurl}


\bibitem[\protect\citeauthoryear{??}{can}{[n. d.]b}]%
        {candle-ecp}
 \bibinfo{year}{[n. d.]}\natexlab{b}.
\newblock \bibinfo{title}{Exascale Deep Learning and Simulation Enabled
  Precision Medicine for Cancer}.
\newblock
\newblock
\urldef\tempurl%
\url{https://candle.cels.anl.gov}
\showURL{%
\tempurl}


\bibitem[\protect\citeauthoryear{??}{nas}{[n. d.]}]%
        {nasonlinebib2019}
 \bibinfo{year}{[n. d.]}\natexlab{}.
\newblock \bibinfo{title}{Literature on Neural Architecture Search}.
\newblock
\newblock
\urldef\tempurl%
\url{https://www.ml4aad.org/automl/literature-on-neural-architecture-search/}
\showURL{%
\tempurl}


\bibitem[\protect\citeauthoryear{??}{nni}{[n. d.]}]%
        {nni2019}
 \bibinfo{year}{[n. d.]}\natexlab{}.
\newblock \bibinfo{title}{Neural Network Intelligence}.
\newblock
\newblock
\urldef\tempurl%
\url{https://github.com/Microsoft/nni}
\showURL{%
\tempurl}


\bibitem[\protect\citeauthoryear{??}{nt3}{[n. d.]}]%
        {nt3-code}
 \bibinfo{year}{[n. d.]}\natexlab{}.
\newblock \bibinfo{title}{{NT3 Benchmark}}.
\newblock
\newblock
\urldef\tempurl%
\url{https://github.com/ECP-CANDLE/Benchmarks/tree/master/Pilot1/NT3}
\showURL{%
\tempurl}


\bibitem[\protect\citeauthoryear{??}{uno}{[n. d.]}]%
        {uno-code}
 \bibinfo{year}{[n. d.]}\natexlab{}.
\newblock \bibinfo{title}{{Uno Benchmark}}.
\newblock
\newblock
\urldef\tempurl%
\url{https://github.com/ECP-CANDLE/Benchmarks/tree/master/Pilot1/Uno}
\showURL{%
\tempurl}


\bibitem[\protect\citeauthoryear{??}{Can}{[n. d.]}]%
        {CancerStat}
 \bibinfo{year}{[n. d.]}\natexlab{}.
\newblock \bibinfo{title}{World Health Organization: Cancer key facts}.
\newblock
\newblock
\urldef\tempurl%
\url{https://www.who.int/news-room/fact-sheets/detail/cancer}
\showURL{%
\tempurl}


\bibitem[\protect\citeauthoryear{Abadi, Barham, Chen, Chen, Davis, Dean, Devin,
  Ghemawat, Irving, Isard, et~al\mbox{.}}{Abadi et~al\mbox{.}}{2016}]%
        {abadi2016tensorflow}
\bibfield{author}{\bibinfo{person}{Mart{\'\i}n Abadi}, \bibinfo{person}{Paul
  Barham}, \bibinfo{person}{Jianmin Chen}, \bibinfo{person}{Zhifeng Chen},
  \bibinfo{person}{Andy Davis}, \bibinfo{person}{Jeffrey Dean},
  \bibinfo{person}{Matthieu Devin}, \bibinfo{person}{Sanjay Ghemawat},
  \bibinfo{person}{Geoffrey Irving}, \bibinfo{person}{Michael Isard},
  {et~al\mbox{.}}} \bibinfo{year}{2016}\natexlab{}.
\newblock \showarticletitle{{TensorFlow: A} system for large-scale machine
  learning}. In \bibinfo{booktitle}{\emph{OSDI}}, Vol.~\bibinfo{volume}{16}.
  \bibinfo{pages}{265--283}.
\newblock


\bibitem[\protect\citeauthoryear{Ashok, Rhinehart, Beainy, and Kitani}{Ashok
  et~al\mbox{.}}{2017}]%
        {ashok2017n2n}
\bibfield{author}{\bibinfo{person}{Anubhav Ashok}, \bibinfo{person}{Nicholas
  Rhinehart}, \bibinfo{person}{Fares Beainy}, {and} \bibinfo{person}{Kris~M
  Kitani}.} \bibinfo{year}{2017}\natexlab{}.
\newblock \showarticletitle{{N2n learning: Network} to network compression via
  policy gradient reinforcement learning}.
\newblock \bibinfo{journal}{\emph{arXiv preprint 1709.06030}}
  (\bibinfo{year}{2017}).
\newblock


\bibitem[\protect\citeauthoryear{Baker, Gupta, Naik, and Raskar}{Baker
  et~al\mbox{.}}{2016}]%
        {baker2016designing}
\bibfield{author}{\bibinfo{person}{Bowen Baker}, \bibinfo{person}{Otkrist
  Gupta}, \bibinfo{person}{Nikhil Naik}, {and} \bibinfo{person}{Ramesh
  Raskar}.} \bibinfo{year}{2016}\natexlab{}.
\newblock \showarticletitle{Designing neural network architectures using
  reinforcement learning}.
\newblock \bibinfo{journal}{\emph{arXiv preprint 1611.02167}}
  (\bibinfo{year}{2016}).
\newblock


\bibitem[\protect\citeauthoryear{Balaprakash, Egele, Salim, Vishwanath, and
  Wild}{Balaprakash et~al\mbox{.}}{2018a}]%
        {deephyper-soft2018}
\bibfield{author}{\bibinfo{person}{P. Balaprakash}, \bibinfo{person}{R. Egele},
  \bibinfo{person}{M. Salim}, \bibinfo{person}{V. Vishwanath}, {and}
  \bibinfo{person}{S.~M. Wild}.} \bibinfo{year}{2018}\natexlab{a}.
\newblock \bibinfo{title}{{DeepHyper: Scalable automated machine learning
  package}}.
\newblock
\newblock
\urldef\tempurl%
\url{https://github.com/deephyper/deephyper}
\showURL{%
\tempurl}


\bibitem[\protect\citeauthoryear{Balaprakash, Salim, Uram, Vishwanath, and
  Wild}{Balaprakash et~al\mbox{.}}{2018b}]%
        {deephyper}
\bibfield{author}{\bibinfo{person}{Prasanna Balaprakash},
  \bibinfo{person}{Michael Salim}, \bibinfo{person}{Thomas Uram},
  \bibinfo{person}{Venkat Vishwanath}, {and} \bibinfo{person}{Stefan Wild}.}
  \bibinfo{year}{2018}\natexlab{b}.
\newblock \showarticletitle{{DeepHyper: A}synchronous Hyperparameter Search for
  Deep Neural Networks}. In \bibinfo{booktitle}{\emph{HiPC 2018: 25th edition
  of the IEEE International Conference on High Performance Computing, Data, and
  Analytics}}.
\newblock


\bibitem[\protect\citeauthoryear{Bello, Zoph, Vasudevan, and Le}{Bello
  et~al\mbox{.}}{2017}]%
        {bello2017neural}
\bibfield{author}{\bibinfo{person}{Irwan Bello}, \bibinfo{person}{Barret Zoph},
  \bibinfo{person}{Vijay Vasudevan}, {and} \bibinfo{person}{Quoc~V Le}.}
  \bibinfo{year}{2017}\natexlab{}.
\newblock \showarticletitle{Neural optimizer search with reinforcement
  learning}. In \bibinfo{booktitle}{\emph{Proceedings of the 34th International
  Conference on Machine Learning}}, Vol.~\bibinfo{volume}{70}. JMLR. org,
  \bibinfo{pages}{459--468}.
\newblock


\bibitem[\protect\citeauthoryear{Ben-Nun and Hoefler}{Ben-Nun and
  Hoefler}{2018}]%
        {distdl-preprint}
\bibfield{author}{\bibinfo{person}{T. Ben-Nun} {and} \bibinfo{person}{T.
  Hoefler}.} \bibinfo{year}{2018}\natexlab{}.
\newblock \showarticletitle{{Demystifying Parallel and Distributed Deep
  Learning: {A}n In-Depth Concurrency Analysis}}.
\newblock \bibinfo{journal}{\emph{CoRR}}  \bibinfo{volume}{abs/1802.09941}
  (\bibinfo{date}{Feb.} \bibinfo{year}{2018}).
\newblock


\bibitem[\protect\citeauthoryear{Bender, Kindermans, Zoph, Vasudevan, and
  Le}{Bender et~al\mbox{.}}{2018a}]%
        {bender2018understanding}
\bibfield{author}{\bibinfo{person}{Gabriel Bender}, \bibinfo{person}{Pieter-Jan
  Kindermans}, \bibinfo{person}{Barret Zoph}, \bibinfo{person}{Vijay
  Vasudevan}, {and} \bibinfo{person}{Quoc Le}.}
  \bibinfo{year}{2018}\natexlab{a}.
\newblock \showarticletitle{Understanding and simplifying one-shot architecture
  search}. In \bibinfo{booktitle}{\emph{International Conference on Machine
  Learning}}. \bibinfo{pages}{549--558}.
\newblock


\bibitem[\protect\citeauthoryear{Bender, Kindermans, Zoph, Vasudevan, and
  Le}{Bender et~al\mbox{.}}{2018b}]%
        {pmlr-v80-bender18a}
\bibfield{author}{\bibinfo{person}{Gabriel Bender}, \bibinfo{person}{Pieter-Jan
  Kindermans}, \bibinfo{person}{Barret Zoph}, \bibinfo{person}{Vijay
  Vasudevan}, {and} \bibinfo{person}{Quoc Le}.}
  \bibinfo{year}{2018}\natexlab{b}.
\newblock \showarticletitle{Understanding and Simplifying One-Shot Architecture
  Search}. In \bibinfo{booktitle}{\emph{Proceedings of the 35th International
  Conference on Machine Learning}} \emph{(\bibinfo{series}{Proceedings of
  Machine Learning Research})}, \bibfield{editor}{\bibinfo{person}{Jennifer Dy}
  {and} \bibinfo{person}{Andreas Krause}} (Eds.), Vol.~\bibinfo{volume}{80}.
  \bibinfo{publisher}{PMLR}, \bibinfo{pages}{550--559}.
\newblock


\bibitem[\protect\citeauthoryear{Bergstra and Bengio}{Bergstra and
  Bengio}{2012}]%
        {bergstra2012random}
\bibfield{author}{\bibinfo{person}{James Bergstra} {and}
  \bibinfo{person}{Yoshua Bengio}.} \bibinfo{year}{2012}\natexlab{}.
\newblock \showarticletitle{Random search for hyper-parameter optimization}.
\newblock \bibinfo{journal}{\emph{Journal of Machine Learning Research}}
  \bibinfo{volume}{13}, \bibinfo{number}{Feb} (\bibinfo{year}{2012}),
  \bibinfo{pages}{281--305}.
\newblock


\bibitem[\protect\citeauthoryear{Bergstra, Yamins, and Cox}{Bergstra
  et~al\mbox{.}}{2013a}]%
        {bergstra2013hyperopt}
\bibfield{author}{\bibinfo{person}{James Bergstra}, \bibinfo{person}{Dan
  Yamins}, {and} \bibinfo{person}{David~D Cox}.}
  \bibinfo{year}{2013}\natexlab{a}.
\newblock \showarticletitle{{Hyperopt: A Python library for optimizing the
  hyperparameters of machine learning algorithms}}. In
  \bibinfo{booktitle}{\emph{Proceedings of the 12th Python in Science
  Conference}}. \bibinfo{pages}{13--20}.
\newblock


\bibitem[\protect\citeauthoryear{Bergstra, Yamins, and Cox}{Bergstra
  et~al\mbox{.}}{2013b}]%
        {bergstra2013making}
\bibfield{author}{\bibinfo{person}{James Bergstra}, \bibinfo{person}{Daniel
  Yamins}, {and} \bibinfo{person}{David~Daniel Cox}.}
  \bibinfo{year}{2013}\natexlab{b}.
\newblock \showarticletitle{Making a science of model search: {H}perparameter
  optimization in hundreds of dimensions for vision architectures}.
\newblock  (\bibinfo{year}{2013}).
\newblock


\bibitem[\protect\citeauthoryear{Brockman, Cheung, Pettersson, Schneider,
  Schulman, Tang, and Zaremba}{Brockman et~al\mbox{.}}{2016}]%
        {openaigym}
\bibfield{author}{\bibinfo{person}{Greg Brockman}, \bibinfo{person}{Vicki
  Cheung}, \bibinfo{person}{Ludwig Pettersson}, \bibinfo{person}{Jonas
  Schneider}, \bibinfo{person}{John Schulman}, \bibinfo{person}{Jie Tang},
  {and} \bibinfo{person}{Wojciech Zaremba}.} \bibinfo{year}{2016}\natexlab{}.
\newblock \bibinfo{title}{{OpenAI} Gym}.
\newblock
\newblock
\showeprint{arXiv:1606.01540}


\bibitem[\protect\citeauthoryear{Bronstein, Bruna, LeCun, Szlam, and
  Vandergheynst}{Bronstein et~al\mbox{.}}{2017}]%
        {bronstein2017geometric}
\bibfield{author}{\bibinfo{person}{Michael~M Bronstein}, \bibinfo{person}{Joan
  Bruna}, \bibinfo{person}{Yann LeCun}, \bibinfo{person}{Arthur Szlam}, {and}
  \bibinfo{person}{Pierre Vandergheynst}.} \bibinfo{year}{2017}\natexlab{}.
\newblock \showarticletitle{Geometric deep learning: {G}oing beyond euclidean
  data}.
\newblock \bibinfo{journal}{\emph{IEEE Signal Processing Magazine}}
  \bibinfo{volume}{34}, \bibinfo{number}{4} (\bibinfo{year}{2017}),
  \bibinfo{pages}{18--42}.
\newblock


\bibitem[\protect\citeauthoryear{Chen, Zhang, Huang, Mu, Meng, and Wang}{Chen
  et~al\mbox{.}}{2018}]%
        {chen2018reinforced}
\bibfield{author}{\bibinfo{person}{Yukang Chen}, \bibinfo{person}{Qian Zhang},
  \bibinfo{person}{Chang Huang}, \bibinfo{person}{Lisen Mu},
  \bibinfo{person}{Gaofeng Meng}, {and} \bibinfo{person}{Xinggang Wang}.}
  \bibinfo{year}{2018}\natexlab{}.
\newblock \showarticletitle{Reinforced Evolutionary Neural Architecture
  Search}.
\newblock \bibinfo{journal}{\emph{arXiv preprint 1808.00193}}
  (\bibinfo{year}{2018}).
\newblock


\bibitem[\protect\citeauthoryear{Chollet et~al\mbox{.}}{Chollet
  et~al\mbox{.}}{2017}]%
        {chollet2017keras}
\bibfield{author}{\bibinfo{person}{Fran{\c{c}}ois Chollet} {et~al\mbox{.}}}
  \bibinfo{year}{2017}\natexlab{}.
\newblock \bibinfo{title}{Keras (2015)}.
\newblock
\newblock


\bibitem[\protect\citeauthoryear{Chrabaszcz, Loshchilov, and Hutter}{Chrabaszcz
  et~al\mbox{.}}{2017}]%
        {chrabaszcz2017downsampled}
\bibfield{author}{\bibinfo{person}{Patryk Chrabaszcz}, \bibinfo{person}{Ilya
  Loshchilov}, {and} \bibinfo{person}{Frank Hutter}.}
  \bibinfo{year}{2017}\natexlab{}.
\newblock \showarticletitle{A downsampled variant of {ImageNet} as an
  alternative to the {CIFAR} datasets}.
\newblock \bibinfo{journal}{\emph{arXiv preprint 1707.08819}}
  (\bibinfo{year}{2017}).
\newblock


\bibitem[\protect\citeauthoryear{Chu, Zhang, Ma, Xu, Li, and Li}{Chu
  et~al\mbox{.}}{2019}]%
        {chu2019fast}
\bibfield{author}{\bibinfo{person}{Xiangxiang Chu}, \bibinfo{person}{Bo Zhang},
  \bibinfo{person}{Hailong Ma}, \bibinfo{person}{Ruijun Xu},
  \bibinfo{person}{Jixiang Li}, {and} \bibinfo{person}{Qingyuan Li}.}
  \bibinfo{year}{2019}\natexlab{}.
\newblock \showarticletitle{Fast, Accurate and Lightweight Super-Resolution
  with Neural Architecture Search}.
\newblock \bibinfo{journal}{\emph{arXiv preprint 1901.07261}}
  (\bibinfo{year}{2019}).
\newblock


\bibitem[\protect\citeauthoryear{Dhariwal, Hesse, Klimov, Nichol, Plappert,
  Radford, Schulman, Sidor, Wu, and Zhokhov}{Dhariwal et~al\mbox{.}}{2017}]%
        {baselines}
\bibfield{author}{\bibinfo{person}{Prafulla Dhariwal},
  \bibinfo{person}{Christopher Hesse}, \bibinfo{person}{Oleg Klimov},
  \bibinfo{person}{Alex Nichol}, \bibinfo{person}{Matthias Plappert},
  \bibinfo{person}{Alec Radford}, \bibinfo{person}{John Schulman},
  \bibinfo{person}{Szymon Sidor}, \bibinfo{person}{Yuhuai Wu}, {and}
  \bibinfo{person}{Peter Zhokhov}.} \bibinfo{year}{2017}\natexlab{}.
\newblock \bibinfo{title}{{OpenAI baselines}}.
\newblock \bibinfo{howpublished}{\url{https://github.com/openai/baselines}}.
\newblock


\bibitem[\protect\citeauthoryear{Dikov, van~der Smagt, and Bayer}{Dikov
  et~al\mbox{.}}{2019}]%
        {dikov2019bayesian}
\bibfield{author}{\bibinfo{person}{Georgi Dikov}, \bibinfo{person}{Patrick
  van~der Smagt}, {and} \bibinfo{person}{Justin Bayer}.}
  \bibinfo{year}{2019}\natexlab{}.
\newblock \showarticletitle{Bayesian Learning of Neural Network Architectures}.
\newblock \bibinfo{journal}{\emph{arXiv preprint 1901.04436}}
  (\bibinfo{year}{2019}).
\newblock


\bibitem[\protect\citeauthoryear{Dixon, Xu, Dileep, Zhan, Song, Le,
  Yard{\i}mc{\i}, Chakraborty, Bann, Wang, et~al\mbox{.}}{Dixon
  et~al\mbox{.}}{2018}]%
        {dixon2018integrative}
\bibfield{author}{\bibinfo{person}{Jesse~R Dixon}, \bibinfo{person}{Jie Xu},
  \bibinfo{person}{Vishnu Dileep}, \bibinfo{person}{Ye Zhan},
  \bibinfo{person}{Fan Song}, \bibinfo{person}{Victoria~T Le},
  \bibinfo{person}{Galip~G{\"u}rkan Yard{\i}mc{\i}}, \bibinfo{person}{Abhijit
  Chakraborty}, \bibinfo{person}{Darrin~V Bann}, \bibinfo{person}{Yanli Wang},
  {et~al\mbox{.}}} \bibinfo{year}{2018}\natexlab{}.
\newblock \showarticletitle{Integrative detection and analysis of structural
  variation in cancer genomes}.
\newblock \bibinfo{journal}{\emph{Nature Genetics}} \bibinfo{volume}{50},
  \bibinfo{number}{10} (\bibinfo{year}{2018}), \bibinfo{pages}{1388}.
\newblock


\bibitem[\protect\citeauthoryear{Elsken, Metzen, and Hutter}{Elsken
  et~al\mbox{.}}{2018}]%
        {elsken2018neural}
\bibfield{author}{\bibinfo{person}{Thomas Elsken}, \bibinfo{person}{Jan~Hendrik
  Metzen}, {and} \bibinfo{person}{Frank Hutter}.}
  \bibinfo{year}{2018}\natexlab{}.
\newblock \showarticletitle{Neural architecture search: {A} survey}.
\newblock \bibinfo{journal}{\emph{arXiv preprint 1808.05377}}
  (\bibinfo{year}{2018}).
\newblock


\bibitem[\protect\citeauthoryear{Fern{\'a}ndez-Delgado, Cernadas, Barro, and
  Amorim}{Fern{\'a}ndez-Delgado et~al\mbox{.}}{2014}]%
        {fernandez2014we}
\bibfield{author}{\bibinfo{person}{Manuel Fern{\'a}ndez-Delgado},
  \bibinfo{person}{Eva Cernadas}, \bibinfo{person}{Sen{\'e}n Barro}, {and}
  \bibinfo{person}{Dinani Amorim}.} \bibinfo{year}{2014}\natexlab{}.
\newblock \showarticletitle{Do we need hundreds of classifiers to solve real
  world classification problems?}
\newblock \bibinfo{journal}{\emph{The Journal of Machine Learning Research}}
  \bibinfo{volume}{15}, \bibinfo{number}{1} (\bibinfo{year}{2014}),
  \bibinfo{pages}{3133--3181}.
\newblock


\bibitem[\protect\citeauthoryear{Floreano, D{\"u}rr, and Mattiussi}{Floreano
  et~al\mbox{.}}{2008}]%
        {floreano2008neuroevolution}
\bibfield{author}{\bibinfo{person}{Dario Floreano}, \bibinfo{person}{Peter
  D{\"u}rr}, {and} \bibinfo{person}{Claudio Mattiussi}.}
  \bibinfo{year}{2008}\natexlab{}.
\newblock \showarticletitle{Neuroevolution: {F}rom architectures to learning}.
\newblock \bibinfo{journal}{\emph{Evolutionary Intelligence}}
  \bibinfo{volume}{1}, \bibinfo{number}{1} (\bibinfo{year}{2008}),
  \bibinfo{pages}{47--62}.
\newblock


\bibitem[\protect\citeauthoryear{Grondman, Busoniu, Lopes, and
  Babuska}{Grondman et~al\mbox{.}}{2012}]%
        {grondman2012survey}
\bibfield{author}{\bibinfo{person}{Ivo Grondman}, \bibinfo{person}{Lucian
  Busoniu}, \bibinfo{person}{Gabriel~AD Lopes}, {and} \bibinfo{person}{Robert
  Babuska}.} \bibinfo{year}{2012}\natexlab{}.
\newblock \showarticletitle{A survey of actor-critic reinforcement learning:
  S{S}tandard and natural policy gradients}.
\newblock \bibinfo{journal}{\emph{IEEE Transactions on Systems, Man, and
  Cybernetics, Part C (Applications and Reviews)}} \bibinfo{volume}{42},
  \bibinfo{number}{6} (\bibinfo{year}{2012}), \bibinfo{pages}{1291--1307}.
\newblock


\bibitem[\protect\citeauthoryear{Guo, Zhong, Wu, Lin, and Yan}{Guo
  et~al\mbox{.}}{2018}]%
        {guo2018irlas}
\bibfield{author}{\bibinfo{person}{Minghao Guo}, \bibinfo{person}{Zhao Zhong},
  \bibinfo{person}{Wei Wu}, \bibinfo{person}{Dahua Lin}, {and}
  \bibinfo{person}{Junjie Yan}.} \bibinfo{year}{2018}\natexlab{}.
\newblock \showarticletitle{{IRLAS: Inverse} Reinforcement Learning for
  Architecture Search}.
\newblock \bibinfo{journal}{\emph{arXiv preprint 1812.05285}}
  (\bibinfo{year}{2018}).
\newblock


\bibitem[\protect\citeauthoryear{Hutter, Kotthoff, and Vanschoren}{Hutter
  et~al\mbox{.}}{2019}]%
        {AutoMLBook2019}
\bibfield{editor}{\bibinfo{person}{F. Hutter}, \bibinfo{person}{L. Kotthoff},
  {and} \bibinfo{person}{J. Vanschoren}} (Eds.).
  \bibinfo{year}{2019}\natexlab{}.
\newblock \bibinfo{booktitle}{\emph{{Automated Machine Learning: Methods,
  Systems, Challenges}}}.
\newblock \bibinfo{publisher}{Springer International Publishing}.
\newblock


\bibitem[\protect\citeauthoryear{Jaderberg, Dalibard, Osindero, Czarnecki,
  Donahue, Razavi, Vinyals, Green, Dunning, Simonyan, et~al\mbox{.}}{Jaderberg
  et~al\mbox{.}}{2017}]%
        {jaderberg2017population}
\bibfield{author}{\bibinfo{person}{Max Jaderberg}, \bibinfo{person}{Valentin
  Dalibard}, \bibinfo{person}{Simon Osindero}, \bibinfo{person}{Wojciech~M
  Czarnecki}, \bibinfo{person}{Jeff Donahue}, \bibinfo{person}{Ali Razavi},
  \bibinfo{person}{Oriol Vinyals}, \bibinfo{person}{Tim Green},
  \bibinfo{person}{Iain Dunning}, \bibinfo{person}{Karen Simonyan},
  {et~al\mbox{.}}} \bibinfo{year}{2017}\natexlab{}.
\newblock \showarticletitle{Population Based Training of Neural Networks}.
\newblock \bibinfo{journal}{\emph{arXiv preprint 1711.09846}}
  (\bibinfo{year}{2017}).
\newblock


\bibitem[\protect\citeauthoryear{Jin, Song, and Hu}{Jin et~al\mbox{.}}{2018}]%
        {jin2018efficient}
\bibfield{author}{\bibinfo{person}{Haifeng Jin}, \bibinfo{person}{Qingquan
  Song}, {and} \bibinfo{person}{Xia Hu}.} \bibinfo{year}{2018}\natexlab{}.
\newblock \showarticletitle{Efficient neural architecture search with network
  morphism}.
\newblock \bibinfo{journal}{\emph{arXiv preprint 1806.10282}}
  (\bibinfo{year}{2018}).
\newblock


\bibitem[\protect\citeauthoryear{Kamath, Singh, and Dutta}{Kamath
  et~al\mbox{.}}{[n. d.]}]%
        {kamathamla}
\bibfield{author}{\bibinfo{person}{Purushotham Kamath},
  \bibinfo{person}{Abhishek Singh}, {and} \bibinfo{person}{Debo Dutta}.}
  \bibinfo{year}{[n. d.]}\natexlab{}.
\newblock \showarticletitle{{AMLA: An AutoML frAmework for Neural Network
  Design}}.
\newblock  (\bibinfo{year}{[n. d.]}).
\newblock


\bibitem[\protect\citeauthoryear{Kandasamy, Neiswanger, Schneider, Poczos, and
  Xing}{Kandasamy et~al\mbox{.}}{2018}]%
        {kandasamy2018neural}
\bibfield{author}{\bibinfo{person}{Kirthevasan Kandasamy},
  \bibinfo{person}{Willie Neiswanger}, \bibinfo{person}{Jeff Schneider},
  \bibinfo{person}{Barnabas Poczos}, {and} \bibinfo{person}{Eric~P Xing}.}
  \bibinfo{year}{2018}\natexlab{}.
\newblock \showarticletitle{Neural architecture search with {Bayesian}
  optimisation and optimal transport}. In \bibinfo{booktitle}{\emph{Advances in
  Neural Information Processing Systems}}. \bibinfo{pages}{2020--2029}.
\newblock


\bibitem[\protect\citeauthoryear{Klein, Falkner, Mansur, and Hutter}{Klein
  et~al\mbox{.}}{2017}]%
        {klein-bayesopt17}
\bibfield{author}{\bibinfo{person}{A. Klein}, \bibinfo{person}{S. Falkner},
  \bibinfo{person}{N. Mansur}, {and} \bibinfo{person}{F. Hutter}.}
  \bibinfo{year}{2017}\natexlab{}.
\newblock \showarticletitle{{RoBO: A flexible and robust Bayesian Optimization
  framework in Python}}. In \bibinfo{booktitle}{\emph{NeurIPS 2017 Bayesian
  Optimization Workshop}}.
\newblock


\bibitem[\protect\citeauthoryear{Klein, Falkner, Springenberg, and
  Hutter}{Klein et~al\mbox{.}}{2016}]%
        {klein2016learning}
\bibfield{author}{\bibinfo{person}{Aaron Klein}, \bibinfo{person}{Stefan
  Falkner}, \bibinfo{person}{Jost~Tobias Springenberg}, {and}
  \bibinfo{person}{Frank Hutter}.} \bibinfo{year}{2016}\natexlab{}.
\newblock \showarticletitle{Learning curve prediction with {Bayesian} neural
  networks}.
\newblock  (\bibinfo{year}{2016}).
\newblock


\bibitem[\protect\citeauthoryear{Li, Jamieson, DeSalvo, Rostamizadeh, and
  Talwalkar}{Li et~al\mbox{.}}{2016}]%
        {li2016hyperband}
\bibfield{author}{\bibinfo{person}{Lisha Li}, \bibinfo{person}{Kevin Jamieson},
  \bibinfo{person}{Giulia DeSalvo}, \bibinfo{person}{Afshin Rostamizadeh},
  {and} \bibinfo{person}{Ameet Talwalkar}.} \bibinfo{year}{2016}\natexlab{}.
\newblock \showarticletitle{Hyperband: {B}andit-based configuration evaluation
  for hyperparameter optimization}.
\newblock  (\bibinfo{year}{2016}).
\newblock


\bibitem[\protect\citeauthoryear{Li and Talwalkar}{Li and Talwalkar}{2019}]%
        {li2019random}
\bibfield{author}{\bibinfo{person}{Liam Li} {and} \bibinfo{person}{Ameet
  Talwalkar}.} \bibinfo{year}{2019}\natexlab{}.
\newblock \showarticletitle{Random Search and Reproducibility for Neural
  Architecture Search}.
\newblock \bibinfo{journal}{\emph{arXiv preprint 1902.07638}}
  (\bibinfo{year}{2019}).
\newblock


\bibitem[\protect\citeauthoryear{Liang, Meyerson, Hodjat, Fink, Mutch, and
  Miikkulainen}{Liang et~al\mbox{.}}{2019}]%
        {liang2019evolutionary}
\bibfield{author}{\bibinfo{person}{Jason Liang}, \bibinfo{person}{Elliot
  Meyerson}, \bibinfo{person}{Babak Hodjat}, \bibinfo{person}{Dan Fink},
  \bibinfo{person}{Karl Mutch}, {and} \bibinfo{person}{Risto Miikkulainen}.}
  \bibinfo{year}{2019}\natexlab{}.
\newblock \showarticletitle{Evolutionary Neural {AutoML} for Deep Learning}.
\newblock \bibinfo{journal}{\emph{arXiv preprint 1902.06827}}
  (\bibinfo{year}{2019}).
\newblock


\bibitem[\protect\citeauthoryear{Liang, Meyerson, and Miikkulainen}{Liang
  et~al\mbox{.}}{2018}]%
        {liang2018evolutionary}
\bibfield{author}{\bibinfo{person}{Jason Liang}, \bibinfo{person}{Elliot
  Meyerson}, {and} \bibinfo{person}{Risto Miikkulainen}.}
  \bibinfo{year}{2018}\natexlab{}.
\newblock \showarticletitle{Evolutionary architecture search for deep multitask
  networks}. In \bibinfo{booktitle}{\emph{Proceedings of the Genetic and
  Evolutionary Computation Conference}}. ACM, \bibinfo{pages}{466--473}.
\newblock


\bibitem[\protect\citeauthoryear{Ling, Hu, Yang, Yang, Li, Lin, Chen, Dong,
  Cao, Tao, et~al\mbox{.}}{Ling et~al\mbox{.}}{2015}]%
        {ling2015extremely}
\bibfield{author}{\bibinfo{person}{Shaoping Ling}, \bibinfo{person}{Zheng Hu},
  \bibinfo{person}{Zuyu Yang}, \bibinfo{person}{Fang Yang},
  \bibinfo{person}{Yawei Li}, \bibinfo{person}{Pei Lin}, \bibinfo{person}{Ke
  Chen}, \bibinfo{person}{Lili Dong}, \bibinfo{person}{Lihua Cao},
  \bibinfo{person}{Yong Tao}, {et~al\mbox{.}}} \bibinfo{year}{2015}\natexlab{}.
\newblock \showarticletitle{Extremely high genetic diversity in a single tumor
  points to prevalence of non-{D}arwinian cell evolution}.
\newblock \bibinfo{journal}{\emph{Proceedings of the National Academy of
  Sciences}} \bibinfo{volume}{112}, \bibinfo{number}{47}
  (\bibinfo{year}{2015}), \bibinfo{pages}{E6496--E6505}.
\newblock


\bibitem[\protect\citeauthoryear{Liu, Zoph, Shlens, Hua, Li, Fei-Fei, Yuille,
  Huang, and Murphy}{Liu et~al\mbox{.}}{2017}]%
        {liu2017progressive}
\bibfield{author}{\bibinfo{person}{Chenxi Liu}, \bibinfo{person}{Barret Zoph},
  \bibinfo{person}{Jonathon Shlens}, \bibinfo{person}{Wei Hua},
  \bibinfo{person}{Li-Jia Li}, \bibinfo{person}{Li Fei-Fei},
  \bibinfo{person}{Alan Yuille}, \bibinfo{person}{Jonathan Huang}, {and}
  \bibinfo{person}{Kevin Murphy}.} \bibinfo{year}{2017}\natexlab{}.
\newblock \showarticletitle{Progressive neural architecture search}.
\newblock \bibinfo{journal}{\emph{arXiv preprint 1712.00559}}
  (\bibinfo{year}{2017}).
\newblock


\bibitem[\protect\citeauthoryear{Lorenzo and Nalepa}{Lorenzo and
  Nalepa}{2018}]%
        {lorenzo2018memetic}
\bibfield{author}{\bibinfo{person}{Pablo~Ribalta Lorenzo} {and}
  \bibinfo{person}{Jakub Nalepa}.} \bibinfo{year}{2018}\natexlab{}.
\newblock \showarticletitle{Memetic evolution of deep neural networks}. In
  \bibinfo{booktitle}{\emph{Proceedings of the Genetic and Evolutionary
  Computation Conference}}. ACM, \bibinfo{pages}{505--512}.
\newblock


\bibitem[\protect\citeauthoryear{Lorenzo, Nalepa, Ramos, and Pastor}{Lorenzo
  et~al\mbox{.}}{2017}]%
        {lorenzo2017hyper}
\bibfield{author}{\bibinfo{person}{Pablo~Ribalta Lorenzo},
  \bibinfo{person}{Jakub Nalepa}, \bibinfo{person}{Luciano~Sanchez Ramos},
  {and} \bibinfo{person}{Jos{\'e}~Ranilla Pastor}.}
  \bibinfo{year}{2017}\natexlab{}.
\newblock \showarticletitle{Hyper-parameter selection in deep neural networks
  using parallel particle swarm optimization}. In
  \bibinfo{booktitle}{\emph{Proceedings of the Genetic and Evolutionary
  Computation Conference Companion}}. ACM, \bibinfo{pages}{1864--1871}.
\newblock


\bibitem[\protect\citeauthoryear{Luong, Pham, and Manning}{Luong
  et~al\mbox{.}}{2015}]%
        {luong2015effective}
\bibfield{author}{\bibinfo{person}{Minh-Thang Luong}, \bibinfo{person}{Hieu
  Pham}, {and} \bibinfo{person}{Christopher~D Manning}.}
  \bibinfo{year}{2015}\natexlab{}.
\newblock \showarticletitle{Effective approaches to attention-based neural
  machine translation}.
\newblock \bibinfo{journal}{\emph{arXiv preprint 1508.04025}}
  (\bibinfo{year}{2015}).
\newblock


\bibitem[\protect\citeauthoryear{Maziarz, Khorlin, de~Laroussilhe, and
  Gesmundo}{Maziarz et~al\mbox{.}}{2018}]%
        {maziarz2018evolutionary}
\bibfield{author}{\bibinfo{person}{Krzysztof Maziarz}, \bibinfo{person}{Andrey
  Khorlin}, \bibinfo{person}{Quentin de Laroussilhe}, {and}
  \bibinfo{person}{Andrea Gesmundo}.} \bibinfo{year}{2018}\natexlab{}.
\newblock \showarticletitle{Evolutionary-Neural Hybrid Agents for Architecture
  Search}.
\newblock \bibinfo{journal}{\emph{arXiv preprint 1811.09828}}
  (\bibinfo{year}{2018}).
\newblock


\bibitem[\protect\citeauthoryear{Miikkulainen, Liang, Meyerson, Rawal, Fink,
  Francon, Raju, Navruzyan, Duffy, and Hodjat}{Miikkulainen
  et~al\mbox{.}}{2017}]%
        {miikkulainen2017evolving}
\bibfield{author}{\bibinfo{person}{Risto Miikkulainen}, \bibinfo{person}{Jason
  Liang}, \bibinfo{person}{Elliot Meyerson}, \bibinfo{person}{Aditya Rawal},
  \bibinfo{person}{Dan Fink}, \bibinfo{person}{Olivier Francon},
  \bibinfo{person}{Bala Raju}, \bibinfo{person}{Arshak Navruzyan},
  \bibinfo{person}{Nigel Duffy}, {and} \bibinfo{person}{Babak Hodjat}.}
  \bibinfo{year}{2017}\natexlab{}.
\newblock \showarticletitle{Evolving deep neural networks}.
\newblock \bibinfo{journal}{\emph{arXiv preprint 1703.00548}}
  (\bibinfo{year}{2017}).
\newblock


\bibitem[\protect\citeauthoryear{Moritz, Nishihara, Wang, Tumanov, Liaw, Liang,
  Elibol, Yang, Paul, Jordan, et~al\mbox{.}}{Moritz et~al\mbox{.}}{2018}]%
        {moritz2018ray}
\bibfield{author}{\bibinfo{person}{Philipp Moritz}, \bibinfo{person}{Robert
  Nishihara}, \bibinfo{person}{Stephanie Wang}, \bibinfo{person}{Alexey
  Tumanov}, \bibinfo{person}{Richard Liaw}, \bibinfo{person}{Eric Liang},
  \bibinfo{person}{Melih Elibol}, \bibinfo{person}{Zongheng Yang},
  \bibinfo{person}{William Paul}, \bibinfo{person}{Michael~I Jordan},
  {et~al\mbox{.}}} \bibinfo{year}{2018}\natexlab{}.
\newblock \showarticletitle{Ray: {A} Distributed Framework for Emerging
  $\{$AI$\}$ Applications}. In \bibinfo{booktitle}{\emph{13th $\{$USENIX$\}$
  Symposium on Operating Systems Design and Implementation ($\{$OSDI$\}$ 18)}}.
  \bibinfo{pages}{561--577}.
\newblock


\bibitem[\protect\citeauthoryear{Negrinho and Gordon}{Negrinho and
  Gordon}{2017}]%
        {negrinho2017deeparchitect}
\bibfield{author}{\bibinfo{person}{Renato Negrinho} {and}
  \bibinfo{person}{Geoff Gordon}.} \bibinfo{year}{2017}\natexlab{}.
\newblock \showarticletitle{Deeparchitect: {A}utomatically designing and
  training deep architectures}.
\newblock \bibinfo{journal}{\emph{arXiv preprint 1704.08792}}
  (\bibinfo{year}{2017}).
\newblock


\bibitem[\protect\citeauthoryear{Nikolaou, Pavlopoulou, Georgakilas, and
  Kyrodimos}{Nikolaou et~al\mbox{.}}{2018}]%
        {nikolaou2018challenge}
\bibfield{author}{\bibinfo{person}{Michail Nikolaou},
  \bibinfo{person}{Athanasia Pavlopoulou}, \bibinfo{person}{Alexandros~G
  Georgakilas}, {and} \bibinfo{person}{Efthymios Kyrodimos}.}
  \bibinfo{year}{2018}\natexlab{}.
\newblock \showarticletitle{The challenge of drug resistance in cancer
  treatment: {A} current overview}.
\newblock \bibinfo{journal}{\emph{Clinical \& Experimental Metastasis}}
  \bibinfo{volume}{35}, \bibinfo{number}{4} (\bibinfo{year}{2018}),
  \bibinfo{pages}{309--318}.
\newblock


\bibitem[\protect\citeauthoryear{Olson, Bartley, Urbanowicz, and Moore}{Olson
  et~al\mbox{.}}{2016}]%
        {OlsonGECCO2016}
\bibfield{author}{\bibinfo{person}{Randal~S. Olson}, \bibinfo{person}{Nathan
  Bartley}, \bibinfo{person}{Ryan~J. Urbanowicz}, {and}
  \bibinfo{person}{Jason~H. Moore}.} \bibinfo{year}{2016}\natexlab{}.
\newblock \showarticletitle{Evaluation of a Tree-based Pipeline Optimization
  Tool for Automating Data Science}. In \bibinfo{booktitle}{\emph{Proceedings
  of the Genetic and Evolutionary Computation Conference 2016}}
  \emph{(\bibinfo{series}{GECCO '16})}. \bibinfo{publisher}{ACM},
  \bibinfo{address}{New York, NY, USA}, \bibinfo{pages}{485--492}.
\newblock
\showISBNx{978-1-4503-4206-3}
\urldef\tempurl%
\url{https://doi.org/10.1145/2908812.2908918}
\showDOI{\tempurl}


\bibitem[\protect\citeauthoryear{Patton, Johnston, Young, Schuman, March,
  Potok, Rose, Lim, Karnowski, Ziatdinov, et~al\mbox{.}}{Patton
  et~al\mbox{.}}{2018}]%
        {patton2018167}
\bibfield{author}{\bibinfo{person}{Robert~M Patton}, \bibinfo{person}{J~Travis
  Johnston}, \bibinfo{person}{Steven~R Young}, \bibinfo{person}{Catherine~D
  Schuman}, \bibinfo{person}{Don~D March}, \bibinfo{person}{Thomas~E Potok},
  \bibinfo{person}{Derek~C Rose}, \bibinfo{person}{Seung-Hwan Lim},
  \bibinfo{person}{Thomas~P Karnowski}, \bibinfo{person}{Maxim~A Ziatdinov},
  {et~al\mbox{.}}} \bibinfo{year}{2018}\natexlab{}.
\newblock \showarticletitle{{167-PFlops deep learning for electron microscopy:
  From learning physics to atomic manipulation}}. In
  \bibinfo{booktitle}{\emph{Proceedings of the International Conference for
  High Performance Computing, Networking, Storage, and Analysis}}. IEEE Press,
  \bibinfo{pages}{50}.
\newblock


\bibitem[\protect\citeauthoryear{Pedregosa, Varoquaux, Gramfort, Michel,
  Thirion, Grisel, Blondel, Prettenhofer, Weiss, Dubourg, Vanderplas, Passos,
  Cournapeau, Brucher, Perrot, and Duchesnay}{Pedregosa et~al\mbox{.}}{2011}]%
        {scikit-learn}
\bibfield{author}{\bibinfo{person}{F. Pedregosa}, \bibinfo{person}{G.
  Varoquaux}, \bibinfo{person}{A. Gramfort}, \bibinfo{person}{V. Michel},
  \bibinfo{person}{B. Thirion}, \bibinfo{person}{O. Grisel},
  \bibinfo{person}{M. Blondel}, \bibinfo{person}{P. Prettenhofer},
  \bibinfo{person}{R. Weiss}, \bibinfo{person}{V. Dubourg}, \bibinfo{person}{J.
  Vanderplas}, \bibinfo{person}{A. Passos}, \bibinfo{person}{D. Cournapeau},
  \bibinfo{person}{M. Brucher}, \bibinfo{person}{M. Perrot}, {and}
  \bibinfo{person}{E. Duchesnay}.} \bibinfo{year}{2011}\natexlab{}.
\newblock \showarticletitle{{Scikit-learn: Machine} Learning in {P}ython}.
\newblock \bibinfo{journal}{\emph{Journal of Machine Learning Research}}
  \bibinfo{volume}{12} (\bibinfo{year}{2011}), \bibinfo{pages}{2825--2830}.
\newblock


\bibitem[\protect\citeauthoryear{Pham, Guan, Zoph, Le, and Dean}{Pham
  et~al\mbox{.}}{2018}]%
        {pham2018efficient}
\bibfield{author}{\bibinfo{person}{Hieu Pham}, \bibinfo{person}{Melody~Y Guan},
  \bibinfo{person}{Barret Zoph}, \bibinfo{person}{Quoc~V Le}, {and}
  \bibinfo{person}{Jeff Dean}.} \bibinfo{year}{2018}\natexlab{}.
\newblock \showarticletitle{Efficient Neural Architecture Search via Parameter
  Sharing}.
\newblock \bibinfo{journal}{\emph{arXiv preprint 1802.03268}}
  (\bibinfo{year}{2018}).
\newblock


\bibitem[\protect\citeauthoryear{Rawal and Miikkulainen}{Rawal and
  Miikkulainen}{2018}]%
        {rawal2018nodes}
\bibfield{author}{\bibinfo{person}{Aditya Rawal} {and} \bibinfo{person}{Risto
  Miikkulainen}.} \bibinfo{year}{2018}\natexlab{}.
\newblock \showarticletitle{From nodes to networks: {E}volving recurrent neural
  networks}.
\newblock \bibinfo{journal}{\emph{arXiv preprint 1803.04439}}
  (\bibinfo{year}{2018}).
\newblock


\bibitem[\protect\citeauthoryear{Real, Aggarwal, Huang, and Le}{Real
  et~al\mbox{.}}{2018}]%
        {real2018regularized}
\bibfield{author}{\bibinfo{person}{Esteban Real}, \bibinfo{person}{Alok
  Aggarwal}, \bibinfo{person}{Yanping Huang}, {and} \bibinfo{person}{Quoc~V
  Le}.} \bibinfo{year}{2018}\natexlab{}.
\newblock \showarticletitle{Regularized evolution for image classifier
  architecture search}.
\newblock \bibinfo{journal}{\emph{arXiv preprint 1802.01548}}
  (\bibinfo{year}{2018}).
\newblock


\bibitem[\protect\citeauthoryear{Reznik, Luna, Aksoy, Liu, La, Ostrovnaya,
  Creighton, Hakimi, and Sander}{Reznik et~al\mbox{.}}{2018}]%
        {reznik2018landscape}
\bibfield{author}{\bibinfo{person}{Ed Reznik}, \bibinfo{person}{Augustin Luna},
  \bibinfo{person}{B{\"u}lent~Arman Aksoy}, \bibinfo{person}{Eric~Minwei Liu},
  \bibinfo{person}{Konnor La}, \bibinfo{person}{Irina Ostrovnaya},
  \bibinfo{person}{Chad~J Creighton}, \bibinfo{person}{A~Ari Hakimi}, {and}
  \bibinfo{person}{Chris Sander}.} \bibinfo{year}{2018}\natexlab{}.
\newblock \showarticletitle{A landscape of metabolic variation across tumor
  types}.
\newblock \bibinfo{journal}{\emph{Cell Systems}} \bibinfo{volume}{6},
  \bibinfo{number}{3} (\bibinfo{year}{2018}), \bibinfo{pages}{301--313}.
\newblock


\bibitem[\protect\citeauthoryear{Rohekar, Nisimov, Gurwicz, Koren, and
  Novik}{Rohekar et~al\mbox{.}}{2018}]%
        {rohekar2018constructing}
\bibfield{author}{\bibinfo{person}{Raanan~Y Rohekar}, \bibinfo{person}{Shami
  Nisimov}, \bibinfo{person}{Yaniv Gurwicz}, \bibinfo{person}{Guy Koren}, {and}
  \bibinfo{person}{Gal Novik}.} \bibinfo{year}{2018}\natexlab{}.
\newblock \showarticletitle{Constructing Deep Neural Networks by {Bayesian}
  Network Structure Learning}. In \bibinfo{booktitle}{\emph{Advances in Neural
  Information Processing Systems}}. \bibinfo{pages}{3051--3062}.
\newblock


\bibitem[\protect\citeauthoryear{Salim, Uram, Childers, Balaprakash,
  Vishwanath, and Papka}{Salim et~al\mbox{.}}{2018}]%
        {balsam}
\bibfield{author}{\bibinfo{person}{Michael~A. Salim},
  \bibinfo{person}{Thomas~D. Uram}, \bibinfo{person}{Taylor Childers},
  \bibinfo{person}{Prasanna Balaprakash}, \bibinfo{person}{Venkatram
  Vishwanath}, {and} \bibinfo{person}{Michael~E. Papka}.}
  \bibinfo{year}{2018}\natexlab{}.
\newblock \showarticletitle{Balsam: {A}utomated Scheduling and Execution of
  Dynamic, Data-Intensive Workflows}. In \bibinfo{booktitle}{\emph{PyHPC 2018:
  Proceedings of the 8th Workshop on Python for High-Performance and Scientific
  Computing}}.
\newblock


\bibitem[\protect\citeauthoryear{Sanchez-Vega, Mina, Armenia, Chatila, Luna,
  La, Dimitriadoy, Liu, Kantheti, Saghafinia, et~al\mbox{.}}{Sanchez-Vega
  et~al\mbox{.}}{2018}]%
        {sanchez2018oncogenic}
\bibfield{author}{\bibinfo{person}{Francisco Sanchez-Vega},
  \bibinfo{person}{Marco Mina}, \bibinfo{person}{Joshua Armenia},
  \bibinfo{person}{Walid~K Chatila}, \bibinfo{person}{Augustin Luna},
  \bibinfo{person}{Konnor~C La}, \bibinfo{person}{Sofia Dimitriadoy},
  \bibinfo{person}{David~L Liu}, \bibinfo{person}{Havish~S Kantheti},
  \bibinfo{person}{Sadegh Saghafinia}, {et~al\mbox{.}}}
  \bibinfo{year}{2018}\natexlab{}.
\newblock \showarticletitle{Oncogenic signaling pathways in the cancer genome
  atlas}.
\newblock \bibinfo{journal}{\emph{Cell}} \bibinfo{volume}{173},
  \bibinfo{number}{2} (\bibinfo{year}{2018}), \bibinfo{pages}{321--337}.
\newblock


\bibitem[\protect\citeauthoryear{Schulman, Wolski, Dhariwal, Radford, and
  Klimov}{Schulman et~al\mbox{.}}{2017}]%
        {schulman2017proximal}
\bibfield{author}{\bibinfo{person}{John Schulman}, \bibinfo{person}{Filip
  Wolski}, \bibinfo{person}{Prafulla Dhariwal}, \bibinfo{person}{Alec Radford},
  {and} \bibinfo{person}{Oleg Klimov}.} \bibinfo{year}{2017}\natexlab{}.
\newblock \showarticletitle{Proximal policy optimization algorithms}.
\newblock \bibinfo{journal}{\emph{arXiv preprint arXiv:1707.06347}}
  (\bibinfo{year}{2017}).
\newblock


\bibitem[\protect\citeauthoryear{Sciuto, Yu, Jaggi, Musat, and Salzmann}{Sciuto
  et~al\mbox{.}}{2019}]%
        {sciuto2019evaluating}
\bibfield{author}{\bibinfo{person}{Christian Sciuto}, \bibinfo{person}{Kaicheng
  Yu}, \bibinfo{person}{Martin Jaggi}, \bibinfo{person}{Claudiu Musat}, {and}
  \bibinfo{person}{Mathieu Salzmann}.} \bibinfo{year}{2019}\natexlab{}.
\newblock \showarticletitle{Evaluating the Search Phase of Neural Architecture
  Search}.
\newblock \bibinfo{journal}{\emph{arXiv preprint 1902.08142}}
  (\bibinfo{year}{2019}).
\newblock


\bibitem[\protect\citeauthoryear{Snoek, Larochelle, and Adams}{Snoek
  et~al\mbox{.}}{2012}]%
        {snoek2012practical}
\bibfield{author}{\bibinfo{person}{Jasper Snoek}, \bibinfo{person}{Hugo
  Larochelle}, {and} \bibinfo{person}{Ryan~P Adams}.}
  \bibinfo{year}{2012}\natexlab{}.
\newblock \showarticletitle{Practical {Bayesian} optimization of machine
  learning algorithms}. In \bibinfo{booktitle}{\emph{Advances in Neural
  Information Processing Systems}}. \bibinfo{pages}{2951--2959}.
\newblock


\bibitem[\protect\citeauthoryear{Snoek, Rippel, Swersky, Kiros, Satish,
  Sundaram, Patwary, Prabhat, and Adams}{Snoek et~al\mbox{.}}{2015}]%
        {snoek2015scalable}
\bibfield{author}{\bibinfo{person}{Jasper Snoek}, \bibinfo{person}{Oren
  Rippel}, \bibinfo{person}{Kevin Swersky}, \bibinfo{person}{Ryan Kiros},
  \bibinfo{person}{Nadathur Satish}, \bibinfo{person}{Narayanan Sundaram},
  \bibinfo{person}{Mostofa Patwary}, \bibinfo{person}{Mr Prabhat}, {and}
  \bibinfo{person}{Ryan Adams}.} \bibinfo{year}{2015}\natexlab{}.
\newblock \showarticletitle{Scalable {B}ayesian optimization using deep neural
  networks}. In \bibinfo{booktitle}{\emph{International Conference on Machine
  Learning}}. \bibinfo{pages}{2171--2180}.
\newblock


\bibitem[\protect\citeauthoryear{Stanley, Clune, Lehman, and
  Miikkulainen}{Stanley et~al\mbox{.}}{2019}]%
        {stanley2019}
\bibfield{author}{\bibinfo{person}{Kenneth~O. Stanley}, \bibinfo{person}{Jeff
  Clune}, \bibinfo{person}{Joel Lehman}, {and} \bibinfo{person}{Risto
  Miikkulainen}.} \bibinfo{year}{2019}\natexlab{}.
\newblock \showarticletitle{Designing neural networks through neuroevolution}.
\newblock \bibinfo{journal}{\emph{Nature Machine Intelligence}}
  \bibinfo{volume}{1}, \bibinfo{number}{1} (\bibinfo{year}{2019}),
  \bibinfo{pages}{24--35}.
\newblock
\showISBNx{2522-5839}
\urldef\tempurl%
\url{https://doi.org/10.1038/s42256-018-0006-z}
\showDOI{\tempurl}


\bibitem[\protect\citeauthoryear{Stanley, D'Ambrosio, and Gauci}{Stanley
  et~al\mbox{.}}{2009}]%
        {stanley2009hypercube}
\bibfield{author}{\bibinfo{person}{Kenneth~O Stanley}, \bibinfo{person}{David~B
  D'Ambrosio}, {and} \bibinfo{person}{Jason Gauci}.}
  \bibinfo{year}{2009}\natexlab{}.
\newblock \showarticletitle{A hypercube-based encoding for evolving large-scale
  neural networks}.
\newblock \bibinfo{journal}{\emph{Artificial Life}} \bibinfo{volume}{15},
  \bibinfo{number}{2} (\bibinfo{year}{2009}), \bibinfo{pages}{185--212}.
\newblock


\bibitem[\protect\citeauthoryear{Suganuma, Ozay, and Okatani}{Suganuma
  et~al\mbox{.}}{2018}]%
        {suganuma2018exploiting}
\bibfield{author}{\bibinfo{person}{Masanori Suganuma}, \bibinfo{person}{Mete
  Ozay}, {and} \bibinfo{person}{Takayuki Okatani}.}
  \bibinfo{year}{2018}\natexlab{}.
\newblock \showarticletitle{Exploiting the potential of standard convolutional
  autoencoders for image restoration by evolutionary search}.
\newblock \bibinfo{journal}{\emph{arXiv preprint 1803.00370}}
  (\bibinfo{year}{2018}).
\newblock


\bibitem[\protect\citeauthoryear{Suganuma, Shirakawa, and Nagao}{Suganuma
  et~al\mbox{.}}{2017}]%
        {suganuma2017genetic}
\bibfield{author}{\bibinfo{person}{Masanori Suganuma},
  \bibinfo{person}{Shinichi Shirakawa}, {and} \bibinfo{person}{Tomoharu
  Nagao}.} \bibinfo{year}{2017}\natexlab{}.
\newblock \showarticletitle{A genetic programming approach to designing
  convolutional neural network architectures}. In
  \bibinfo{booktitle}{\emph{Proceedings of the Genetic and Evolutionary
  Computation Conference}}. ACM, \bibinfo{pages}{497--504}.
\newblock


\bibitem[\protect\citeauthoryear{Sutton and Barto}{Sutton and Barto}{2018}]%
        {sutton2018reinforcement}
\bibfield{author}{\bibinfo{person}{Richard~S. Sutton} {and}
  \bibinfo{person}{Andrew~G. Barto}.} \bibinfo{year}{2018}\natexlab{}.
\newblock \bibinfo{booktitle}{\emph{Reinforcement learning: An introduction}}.
\newblock \bibinfo{publisher}{MIT Press}.
\newblock


\bibitem[\protect\citeauthoryear{Sutton, McAllester, Singh, and Mansour}{Sutton
  et~al\mbox{.}}{2000}]%
        {sutton2000policy}
\bibfield{author}{\bibinfo{person}{Richard~S. Sutton},
  \bibinfo{person}{David~A. McAllester}, \bibinfo{person}{Satinder~P. Singh},
  {and} \bibinfo{person}{Yishay Mansour}.} \bibinfo{year}{2000}\natexlab{}.
\newblock \showarticletitle{Policy gradient methods for reinforcement learning
  with function approximation}. In \bibinfo{booktitle}{\emph{Advances in neural
  information processing systems}}. \bibinfo{pages}{1057--1063}.
\newblock


\bibitem[\protect\citeauthoryear{van Wyk and Bosman}{van Wyk and
  Bosman}{2018}]%
        {van2018evolutionary}
\bibfield{author}{\bibinfo{person}{Gerard~Jacques van Wyk} {and}
  \bibinfo{person}{Anna~Sergeevna Bosman}.} \bibinfo{year}{2018}\natexlab{}.
\newblock \showarticletitle{Evolutionary Neural Architecture Search for Image
  Restoration}.
\newblock \bibinfo{journal}{\emph{arXiv preprint 1812.05866}}
  (\bibinfo{year}{2018}).
\newblock


\bibitem[\protect\citeauthoryear{Vaswani, Shazeer, Parmar, Uszkoreit, Jones,
  Gomez, Kaiser, and Polosukhin}{Vaswani et~al\mbox{.}}{2017}]%
        {vaswani2017attention}
\bibfield{author}{\bibinfo{person}{Ashish Vaswani}, \bibinfo{person}{Noam
  Shazeer}, \bibinfo{person}{Niki Parmar}, \bibinfo{person}{Jakob Uszkoreit},
  \bibinfo{person}{Llion Jones}, \bibinfo{person}{Aidan~N Gomez},
  \bibinfo{person}{{\L}ukasz Kaiser}, {and} \bibinfo{person}{Illia
  Polosukhin}.} \bibinfo{year}{2017}\natexlab{}.
\newblock \showarticletitle{Attention is all you need}. In
  \bibinfo{booktitle}{\emph{Advances in Neural Information Processing
  Systems}}. \bibinfo{pages}{5998--6008}.
\newblock


\bibitem[\protect\citeauthoryear{Wang, Xu, and Wang}{Wang
  et~al\mbox{.}}{2018}]%
        {wang2018combination}
\bibfield{author}{\bibinfo{person}{Jiazhuo Wang}, \bibinfo{person}{Jason Xu},
  {and} \bibinfo{person}{Xuejun Wang}.} \bibinfo{year}{2018}\natexlab{}.
\newblock \showarticletitle{Combination of hyperband and {Bayesian}
  optimization for hyperparameter optimization in deep learning}.
\newblock \bibinfo{journal}{\emph{arXiv preprint 1801.01596}}
  (\bibinfo{year}{2018}).
\newblock


\bibitem[\protect\citeauthoryear{Wierstra, Gomez, and Schmidhuber}{Wierstra
  et~al\mbox{.}}{2005}]%
        {wierstra2005modeling}
\bibfield{author}{\bibinfo{person}{Daan Wierstra}, \bibinfo{person}{Faustino~J
  Gomez}, {and} \bibinfo{person}{J{\"u}rgen Schmidhuber}.}
  \bibinfo{year}{2005}\natexlab{}.
\newblock \showarticletitle{Modeling systems with internal state using
  {E}volino}. In \bibinfo{booktitle}{\emph{Proceedings of the 7th Annual
  Conference on Genetic and Evolutionary Computation}}. ACM,
  \bibinfo{pages}{1795--1802}.
\newblock


\bibitem[\protect\citeauthoryear{Wistuba}{Wistuba}{2017}]%
        {wistubabayesian}
\bibfield{author}{\bibinfo{person}{Martin Wistuba}.}
  \bibinfo{year}{2017}\natexlab{}.
\newblock \showarticletitle{Bayesian Optimization Combined with Incremental
  Evaluation for Neural Network Architecture Optimization}.
\newblock \bibinfo{journal}{\emph{Proceedings of the International Workshop on
  Automatic Selection, Configuration and Composition of Machine Learning
  Algorithms}} (\bibinfo{year}{2017}).
\newblock


\bibitem[\protect\citeauthoryear{Wozniak, Jain, Balaprakash, Ozik, Collier,
  Bauer, Xia, Brettin, Stevens, Mohd{-}Yusof, Garcia{-}Cardona, Essen, and
  Baughman}{Wozniak et~al\mbox{.}}{2018}]%
        {WozBMC18}
\bibfield{author}{\bibinfo{person}{Justin~M. Wozniak}, \bibinfo{person}{Rajeev
  Jain}, \bibinfo{person}{Prasanna Balaprakash}, \bibinfo{person}{Jonathan
  Ozik}, \bibinfo{person}{Nicholson~T. Collier}, \bibinfo{person}{John Bauer},
  \bibinfo{person}{Fangfang Xia}, \bibinfo{person}{Thomas~S. Brettin},
  \bibinfo{person}{Rick Stevens}, \bibinfo{person}{Jamaludin Mohd{-}Yusof},
  \bibinfo{person}{Cristina Garcia{-}Cardona}, \bibinfo{person}{Brian~Van
  Essen}, {and} \bibinfo{person}{Matthew Baughman}.}
  \bibinfo{year}{2018}\natexlab{}.
\newblock \showarticletitle{{CANDLE/Supervisor: A workflow framework for
  machine learning applied to cancer research}}.
\newblock \bibinfo{journal}{\emph{{BMC} Bioinformatics}}
  \bibinfo{volume}{19-S}, \bibinfo{number}{18} (\bibinfo{year}{2018}),
  \bibinfo{pages}{59--69}.
\newblock
\urldef\tempurl%
\url{https://doi.org/10.1186/s12859-018-2508-4}
\showDOI{\tempurl}


\bibitem[\protect\citeauthoryear{Xia, Shukla, Brettin, Garcia-Cardona, Cohn,
  Allen, Maslov, Holbeck, Doroshow, Evrard, et~al\mbox{.}}{Xia
  et~al\mbox{.}}{2018}]%
        {xia2018predicting}
\bibfield{author}{\bibinfo{person}{Fangfang Xia}, \bibinfo{person}{Maulik
  Shukla}, \bibinfo{person}{Thomas Brettin}, \bibinfo{person}{Cristina
  Garcia-Cardona}, \bibinfo{person}{Judith Cohn}, \bibinfo{person}{Jonathan~E
  Allen}, \bibinfo{person}{Sergei Maslov}, \bibinfo{person}{Susan~L Holbeck},
  \bibinfo{person}{James~H Doroshow}, \bibinfo{person}{Yvonne~A Evrard},
  {et~al\mbox{.}}} \bibinfo{year}{2018}\natexlab{}.
\newblock \showarticletitle{Predicting tumor cell line response to drug pairs
  with deep learning}.
\newblock \bibinfo{journal}{\emph{BMC Bioinformatics}} \bibinfo{volume}{19},
  \bibinfo{number}{18} (\bibinfo{year}{2018}), \bibinfo{pages}{486}.
\newblock


\bibitem[\protect\citeauthoryear{Xie, Zheng, Liu, and Lin}{Xie
  et~al\mbox{.}}{2018}]%
        {xie2018snas}
\bibfield{author}{\bibinfo{person}{Sirui Xie}, \bibinfo{person}{Hehui Zheng},
  \bibinfo{person}{Chunxiao Liu}, {and} \bibinfo{person}{Liang Lin}.}
  \bibinfo{year}{2018}\natexlab{}.
\newblock \showarticletitle{{SNAS: Stochastic neural architecture search}}.
\newblock \bibinfo{journal}{\emph{arXiv preprint 1812.09926}}
  (\bibinfo{year}{2018}).
\newblock


\bibitem[\protect\citeauthoryear{Young, Rose, Johnston, Heller, Karnowski,
  Potok, Patton, Perdue, and Miller}{Young et~al\mbox{.}}{2017}]%
        {young2017evolving}
\bibfield{author}{\bibinfo{person}{Steven~R Young}, \bibinfo{person}{Derek~C
  Rose}, \bibinfo{person}{Travis Johnston}, \bibinfo{person}{William~T Heller},
  \bibinfo{person}{Thomas~P Karnowski}, \bibinfo{person}{Thomas~E Potok},
  \bibinfo{person}{Robert~M Patton}, \bibinfo{person}{Gabriel Perdue}, {and}
  \bibinfo{person}{Jonathan Miller}.} \bibinfo{year}{2017}\natexlab{}.
\newblock \showarticletitle{Evolving deep networks using {HPC}}. In
  \bibinfo{booktitle}{\emph{Proceedings of the Machine Learning on HPC
  Environments}}. ACM.
\newblock


\bibitem[\protect\citeauthoryear{Zela, Klein, Falkner, and Hutter}{Zela
  et~al\mbox{.}}{2018}]%
        {zela2018towards}
\bibfield{author}{\bibinfo{person}{Arber Zela}, \bibinfo{person}{Aaron Klein},
  \bibinfo{person}{Stefan Falkner}, {and} \bibinfo{person}{Frank Hutter}.}
  \bibinfo{year}{2018}\natexlab{}.
\newblock \showarticletitle{Towards automated deep learning: Efficient joint
  neural architecture and hyperparameter search}.
\newblock \bibinfo{journal}{\emph{arXiv preprint 1807.06906}}
  (\bibinfo{year}{2018}).
\newblock


\bibitem[\protect\citeauthoryear{Zoph and Le}{Zoph and Le}{2016}]%
        {zoph2016neural}
\bibfield{author}{\bibinfo{person}{Barret Zoph} {and} \bibinfo{person}{Quoc~V
  Le}.} \bibinfo{year}{2016}\natexlab{}.
\newblock \showarticletitle{Neural architecture search with reinforcement
  learning}.
\newblock \bibinfo{journal}{\emph{arXiv preprint 1611.01578}}
  (\bibinfo{year}{2016}).
\newblock


\bibitem[\protect\citeauthoryear{Zoph, Vasudevan, Shlens, and Le}{Zoph
  et~al\mbox{.}}{2018}]%
        {zoph2018learning}
\bibfield{author}{\bibinfo{person}{Barret Zoph}, \bibinfo{person}{Vijay
  Vasudevan}, \bibinfo{person}{Jonathon Shlens}, {and} \bibinfo{person}{Quoc~V
  Le}.} \bibinfo{year}{2018}\natexlab{}.
\newblock \showarticletitle{Learning transferable architectures for scalable
  image recognition}. In \bibinfo{booktitle}{\emph{Proceedings of the IEEE
  Conference on Computer Vision and Pattern Recognition}}.
  \bibinfo{pages}{8697--8710}.
\newblock


\end{thebibliography}

\begin{center}
    \framebox{\parbox{2.5in}{
    The submitted manuscript has been created by UChicago Argonne, LLC, Operator of Argonne National Laboratory (``Argonne''). Argonne, a U.S. Department of Energy Office of Science laboratory, is operated under Contract No. DE-AC02-06CH11357. The U.S. Government retains for itself, and others acting on its behalf, a paid-up nonexclusive, irrevocable worldwide license in said article to reproduce, prepare derivative works, distribute copies to the public, and perform publicly and display publicly, by or on behalf of the Government. The Department of Energy will provide public access to these results of federally sponsored research in accordance with the DOE Public Access Plan. \url{http://energy.gov/downloads/doe-public-access-plan}}}
    \normalsize
\end{center}

\end{document}